%% file: iclr2027_conference.tex
\documentclass{article} 
\usepackage{iclr2027_conference,times}

\makeatletter
\def\paragraph{\@startsection{paragraph}{4}{\z@}{0.6ex plus
0.3ex minus .1ex}{-1em}{\normalsize\bf}}
\def\section{\@startsection {section}{1}{\z@}{-1.5ex plus
    -0.4ex minus -.2ex}{1.0ex plus 0.2ex
minus0.1ex}{\large\sc\raggedright}}

\def\subsection{\@startsection{subsection}{2}{\z@}{-1.2ex plus
-0.3ex minus -.1ex}{0.6ex plus .1ex}{\normalsize\sc\raggedright}}
\makeatother

\input{math_commands.tex}

\usepackage[pdfborder={0 0 0}]{hyperref}
\usepackage{url}
\usepackage{graphicx}
\usepackage{xcolor}
\usepackage{tcolorbox}
\usepackage{amsthm}
\usepackage{thm-restate}
\usepackage{wrapfig}
\usepackage{booktabs}
\usepackage{multirow}

\newtheorem{theorem}{Theorem}[section]
\newtheorem{proposition}[theorem]{Proposition}

\newtheorem{remark}[theorem]{Remark}
\newtheorem{definition}[theorem]{Definition}

\newenvironment{messagebox}{%
  \begin{tcolorbox}[
    colback=blue!5,
    colframe=blue!70!black,
    arc=4pt,
    boxrule=1.2pt,
    width=\textwidth,
    left=12pt,
    right=12pt,
    top=6pt,
    bottom=6pt
  ]
}{%
  \end{tcolorbox}
}

\title {Understanding the Importance of Massive \\ Activations through Adversarial Robustness}

\author{\textbf{Tigran Fahradyan$^{1}$, 
Riccardo Ali$^{1}$, 
Sergio Calvo-Ordoñez$^{2}$} \\
\textbf{José Miguel Hernández-Lobato$^{1}$,
Pietro Li\`{o}$^{1}$, 
\'{A}lvaro Arroyo$^{2}$} \\
$^{1}$University of Cambridge $^{2}$University of Oxford
}

\iclrfinalcopy 
\begin{document}

\maketitle

\lhead{}

\begin{abstract}
  A common occurrence in Large Language Models (LLMs) is the emergence of so-called \textit{massive activations}. This phenomenon is well known in the literature and has been connected to attention sinks, representational compression, as well as signal propagation. However, the reasons and consequences for their appearance are not yet well understood. In this paper, we explain the role of massive activations in deep learning by showing that they are connected to the \textit{region migration} phenomenon in delayed robustness. Under the spline theoretic interpretation of deep networks, we show that massive activations push the representations away from the nonlinearity boundaries of the network, decreasing the complexity of its input-output map around data points. Furthermore, we uncover for the first time that this phenomenon is not unique to transformers and actually appears in a number of common architectures performing simple supervised learning tasks. We validate this connection through extensive experiments on a diverse set of architectures including residual MLP, ResNet18, Vision Transformers and GPT and datasets including MNIST, CIFAR10, Imagenette and Shakespeare Text. We consistently observe that massive activations and region migration co-occur with adversarial robustness. Furthermore, we illustrate that this connection holds in the large-scale pretraining of LLMs by showing that the rise of massive activations in LLMs closely tracks the adversarial accuracy of the model under perturbations of the token embeddings. Code is available \href{https://github.com/tigranfah/ma-robustness}{\textcolor{blue}{here}}.
\end{abstract}

\section{Introduction}

Despite their remarkable success, the inner workings of neural networks remain poorly understood, giving rise to surprising phenomena that challenge our understanding of how they learn. One such phenomenon that recently gained traction in the context of LLMs is the so-called \textit{massive activations}~\citep{sun2024massive}. These are a few activations in the hidden representation of neural networks that grow multiple orders of magnitude larger than the others. These activations rise in early layers and persist through the network via the residual stream, until disappearing near the output. Recent work has linked massive activations to attention sinks and pre-norm architectures \citep{sun2026spike}, traced their formation to interactions between normalization and feed-forward layers \citep{shi2026single}, and interpreted them as structural bias vectors \citep{chen2026massive}. Related work on activation outliers has also implicated normalization, signal propagation, and optimization in their emergence \citep{he2024understanding}. However, their broader functional role remains poorly understood. 

This work is, to the best of our knowledge, the first to empirically and theoretically show the connection between massive activations, \textit{delayed robustness} and \textit{adversarial grokking} \citep{humayun2024deep} (the emergence of adversarial robustness \citep{Szegedy2014Intriguing, Goodfellow2015harness, madry2018towards} long after generalization). Our central finding is that massive activations, emerging in the late phases of training, simplify the local input-output map of neural networks, inducing the phenomenon called \textit{region migration}~\citep{humayun2024deep} in delayed robustness, which empirically coincides with increased adversarial robustness.

To study this connection in a principled manner, we view neural networks through the lens of Spline Theory of Deep Networks \citep{balestriero2018spline, balestriero2020mad}. Under this interpretation, a neural network partitions its input space into convex polytopal regions and applies an affine mapping on each region. We show that massive activations displace the data points onto large linear regions and away from nonlinearity boundaries. This displacement of representations is the \textit{region migration} phenomenon observed in delayed robustness \citep{humayun2024deep}.

We provide extensive experimental support for this connection, on a diverse mix of architectures, like MLPs \citep{lecun1998mnist} with residual connections, ResNet18 \citep{he2016deep}, Vision Transformers (ViTs) \citep{dosovitskiy2021an} and GPTs and datasets, like MNIST \citep{lecun1998mnist}, CIFAR10 \citep{Krizhevsky09learningmultiple}, Imagenette \citep{howard2019imagenette} and Shakespeare Text \citep{karpathy2015shakespeare}. Across all the experiments we consistently observe that massive activations and region migration co-occur with adversarial robustness, providing strong empirical support for our proposed connection. Furthermore, we investigate this connection in LLMs and show that during pretraining the rise of massive activations closely tracks the onset of region migration and robustness to perturbations in the token embedding space.

We summarize our contributions as follows:

\begin{itemize}
    \item We show that massive activations are more widespread in deep learning than previously known (they do not appear only in transformers, as was originally found in \citet{sun2024massive}).
    \item We provide extensive empirical evidence showing that massive activations cause region migration and co-occur with adversarial robustness, providing a new perspective on the role of massive activations and why they are useful in deep learning.
    \item Under the spline theoretic view of deep networks, we theoretically show that massive activations displace the representations away from nonlinearity boundaries, simplifying the neural network input-output map.
    \item We demonstrate that our finding is also relevant for LLMs, by showing that the massive activations emerging during pretraining induce region migration and closely follow the robustness to perturbations of the token embeddings.
\end{itemize}

\section{Preliminaries and Related Work}
\label{sec:preliminaries}

\subsection{Massive Activations}

\textit{Massive activations} are a surprising empirical phenomenon, observed in the hidden representations of large neural networks \citep{sun2024massive}. \citet{sun2024massive} conduct a systematic study of these activations and report that massive activations have two defining characteristics: first, there are very few of them, typically fewer than ten in the hidden representation; second, they are a few orders of magnitude larger than the other activations.

\citet{sun2024massive} observed that massive activations emerge as early as after the computation of the first layer in the network, remain roughly constant in value throughout most of the intermediate layers and diminish in the final few layers. Massive activations reside in a fixed location of the hidden representations, are typically input independent, and often occur in the beginning of a sequence, period or newline token, which do not have semantic meaning.

\subsubsection{What are Massive Activations For?} \paragraph{Fixed Biases.} \citet{sun2024massive} show that massive activations are not redundant and act as fixed but important biases in the network. Notably, when they are set to zero, the model performance degrades significantly; on the other hand, setting massive activations to their mean value computed across different samples, causes negligible performance change.

\paragraph{Attention Sinks.} \citet{sun2024massive, queipo-de-llano2026attention, sun2026anatomy} claimed that massive activations cause \textit{attention sinks} \citep{xiao2024efficient}. The connection is due to the fact that massive activations appear in the attention key vectors of their associated tokens and the dot product between any query and these key vectors produces a large activation logit. As a result, the large activation logit dominates the softmax operation, since the softmax output must sum to one and nearly all the probability mass is allocated to the token with the massive activation.

\paragraph{Mitigating Over-mixing.} \citet{barbero2025why} show that attention sinks, which are a consequence of massive activations on the first token \citep{sun2024massive, queipo-de-llano2026attention}, are useful in transformers, because they prevent over-mixing of the representations and enable faithful information propagation through the layers, providing an additional interpretation of their role and why they might be useful in deep learning.

\paragraph{Compression Valleys.} \citet{queipo-de-llano2026attention} showed that massive activations compress the hidden representations, which is captured by the low entropy of the hidden representations in the middle layers. Massive activations cause the representation norm of a single token to dominate the others, creating a dominant singular value in the hidden representation matrix. The dominant singular value suppresses the contributions of all the tokens, causing the matrix based entropy of the representation to drop.

\paragraph{Natural Robustness.} The closest work in the literature to ours is \citet{2024interpreting}, which identifies a connection between large activations and natural robustness (robustness to ImageNet \citep{imagenet} distribution shifts). While \citet{2024interpreting} shows that large activations are indicative of natural robustness, our work and contributions differ in the following important ways. Firstly, we consider adversarial robustness that measures robustness under carefully crafted input perturbations that aim to cause misclassification, which meaningfully differs from natural robustness. Additionally, we provide empirical and theoretical connections between massive activations and region migration \citep{humayun2024deep}, which is associated with emergent adversarial robustness. In contrast to \citet{2024interpreting} we validate this connection across a diverse mix of architecture and dataset choices.



\begin{figure}[t]
  \centering
  \begin{minipage}{0.376\textwidth}
      \centering
      \includegraphics[width=\linewidth]{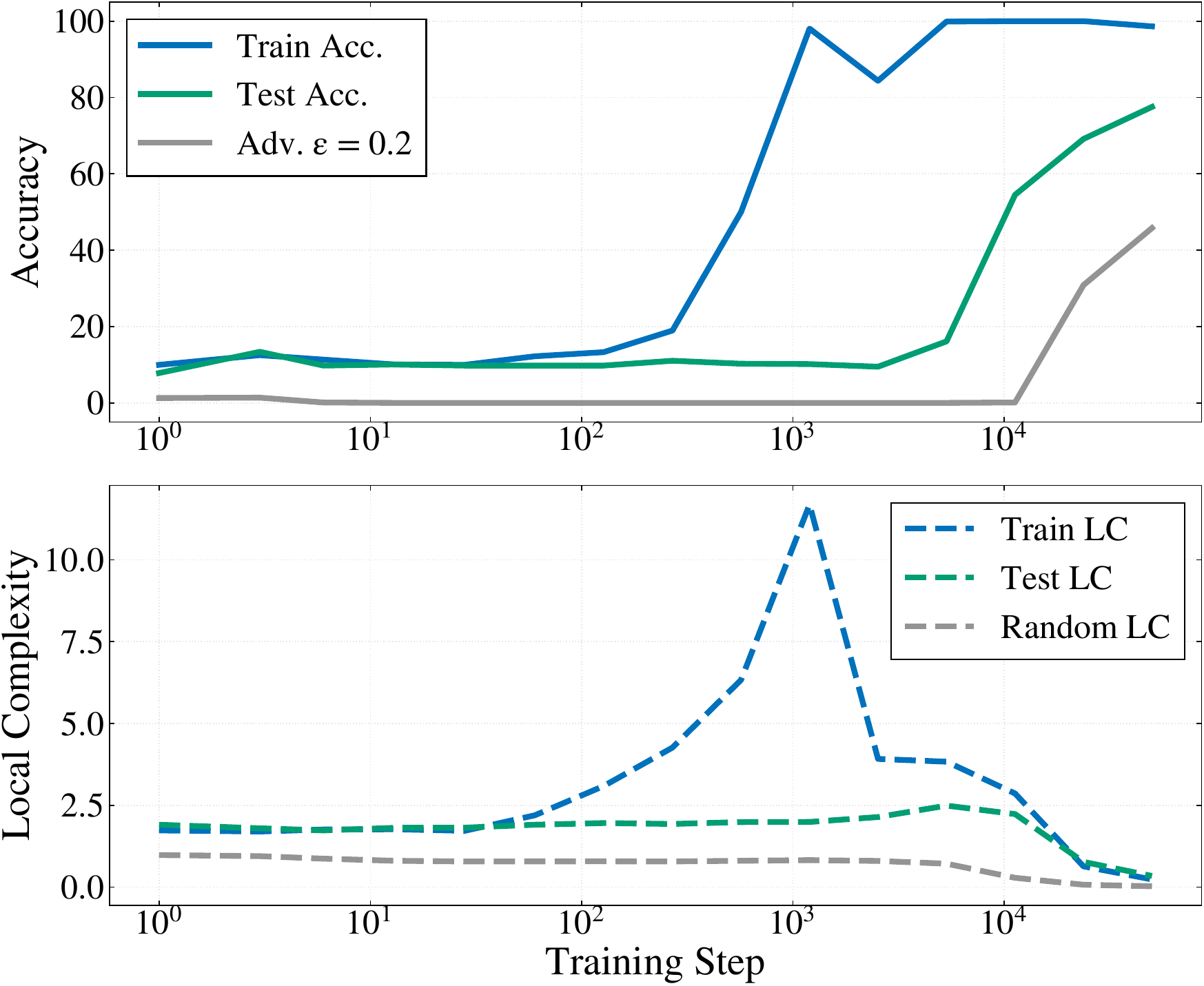}
  \end{minipage}
  \hfill
  \begin{minipage}{0.6\textwidth}
      \centering
      \raisebox{0.34cm}{\includegraphics[width=\linewidth]{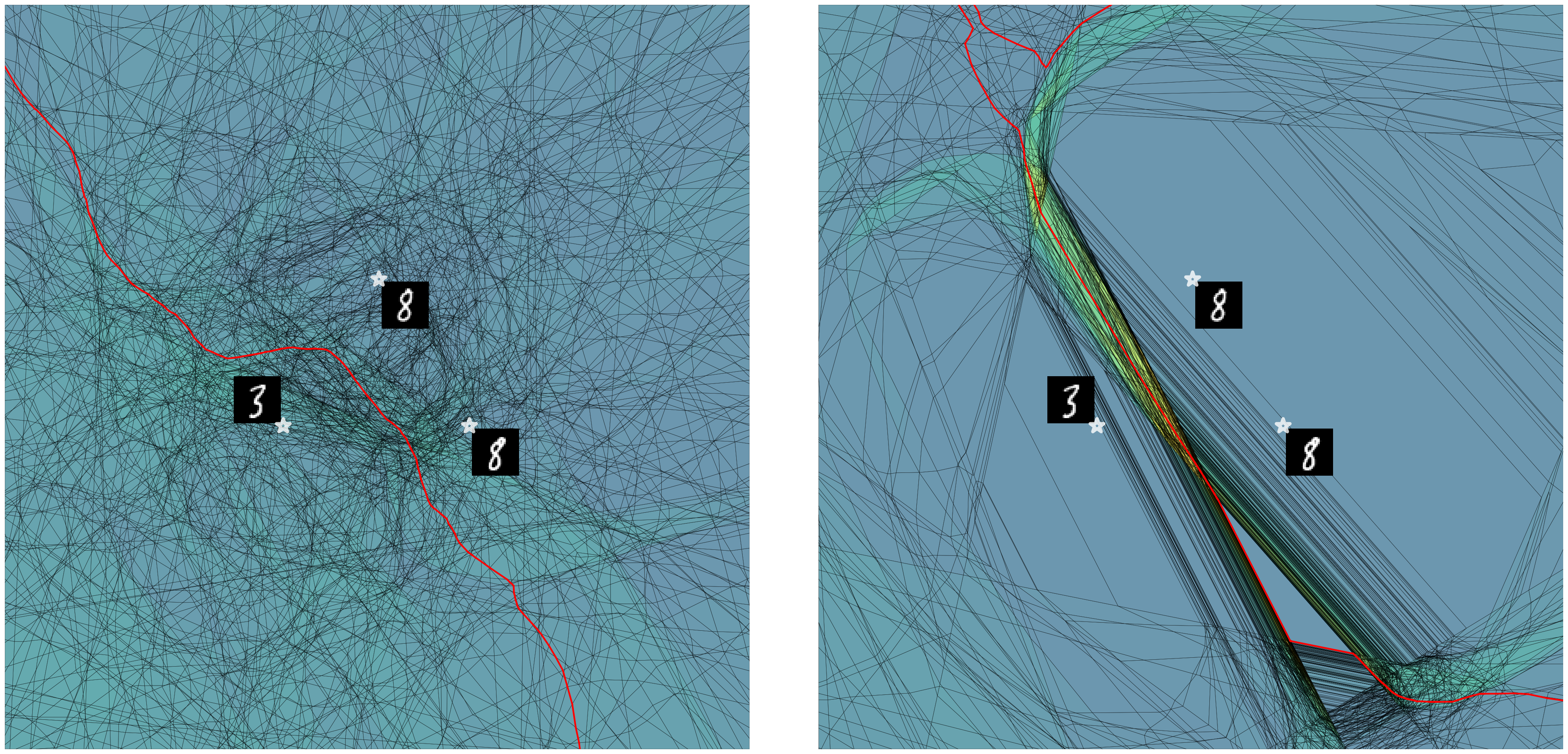}}
  \end{minipage}
  \caption{Residual MLP training on MNIST digit classification dataset. \textbf{Left:} train, test and adversarial accuracies along with the LC on train, test and random points. \textbf{Middle and Right:} SplineCam visualization of the input space partitioning of the residual MLP that passes through three data points. The black lines in the SplineCam plots represent the nonlinearity boundaries of the model, where on each induced cell or region the neural network is an affine map. Stars represent the data points, red line is the decision boundary and colors of the regions correspond to the norm of their affine mappings. We provide more detailed explanation about SplineCam visualization in \ref{app:splinecam}}
  \label{fig:mlp_acc_lc_splinecam}
\end{figure}

\subsection{Delayed Robustness and Region Migration}

\citet{humayun2024deep} found a novel, previously unknown form of grokking \citep{power2022grokking,liu2023omnigrok,kumar2024grokking}, in which adversarial accuracy always exhibits a delayed transition, after the saturation of train and test accuracy, and termed this \textit{delayed robustness}. \citet{humayun2024deep} showed that across a wide range of architectures and training settings, deep networks become adversarially robust long after generalization. This means that there is a third phase beyond memorization and generalization where the network reorganizes its parameters to become adversarially robust to input perturbations, which \citet{humayun2024deep} measures with Projected Gradient Descent (PGD) attack \citep{madry2018towards} with different radii $\epsilon$. This phenomenon has been established to be a general occurrence in deep learning and consistently observed in MLPs trained on MNIST \citep{lecun1998mnist}, CNNs \citep{lecun1998mnist} and ResNets \citep{he2016deep} trained on CIFAR10, CIFAR100 \citep{Krizhevsky09learningmultiple} and Imagenette \citep{howard2019imagenette} and GPT trained on Shakespeare Text \citep{karpathy2015shakespeare}.
\paragraph{Local Complexity.} To track the emergence of delayed robustness, \citet{humayun2024deep} propose a \textit{local complexity} (LC) progress measure \citep{barak2022hidden, nanda2023progress}, which quantifies the density of the nonlinearity boundaries around data points, that serves as a measure of local expressivity. More concretely, LC is computed by first constructing a small cross-polytopal frame around an input point $\vx$. Let $\{\vv_1,\dots,\vv_P\}$ be $P$ orthonormal directions, $r>0$ be a neighborhood radius and $\gV_{\vx}^{(0)} =\{\vx \pm r\vv_p : p=1,\dots,P\}$ be the input cross-polytopal frame. The convex hull of this frame approximates a local neighborhood of the data point. The frame is constructed once in the input space and then propagated through the network layer by layer. At each layer, the metric asks how many neuron nonlinearity boundaries cut through the propagated neighborhood. Operationally, this is checked by evaluating each neuron's preactivation on the vertices of the propagated frame. If different vertices fall on different sides of the same nonlinearity boundary, then the frame has crossed that boundary and the neuron contributes to the layer's LC. The total LC is obtained by summing these boundary-crossing counts over all layers. Thus, LC should be understood geometrically as the number of nonlinear boundaries intersected by a small neighborhood of the data point as that neighborhood travels through the network. A large LC indicates that the neighborhood is cut by many activation boundaries and can move between many affine maps under small perturbations, while a small LC indicates that the propagated neighborhood lies inside larger affine regions.
\vspace{-.5em}
\paragraph{Delayed Robustness.} \citet{humayun2024deep} study the LC throughout the training of deep networks and find that it undergoes a double descent (first decreases, then increases and then decreases). \citet{humayun2024deep} shows that the network starts becoming robust against PGD perturbations during the second descent of LC, also called \textit{region migration}, and this is when it reaches the \textit{delayed robustness} state. We provide an illustration of delayed robustness for MLP with residual connections on MNIST dataset in Figure \ref{fig:mlp_acc_lc_splinecam}. Figure \ref{fig:mlp_acc_lc_splinecam} (left) shows the train, test and adversarial accuracy on test set across steps along with the LC metric on train, test and random points. The adversarial accuracy increases with the decrease of the LC metric after generalization. Figure \ref{fig:mlp_acc_lc_splinecam} (middle and right) shows SplineCam \citep{humayun2023splinecam} visualization of the input space partitioning of the network before region migration (middle), where the nonlinearity boundaries are dense around data points (marked with stars) and after region migration (right), where the nonlinearity boundaries migrated away from data points.


\begin{figure}[t]
\centering
\includegraphics[width=1.0\linewidth]{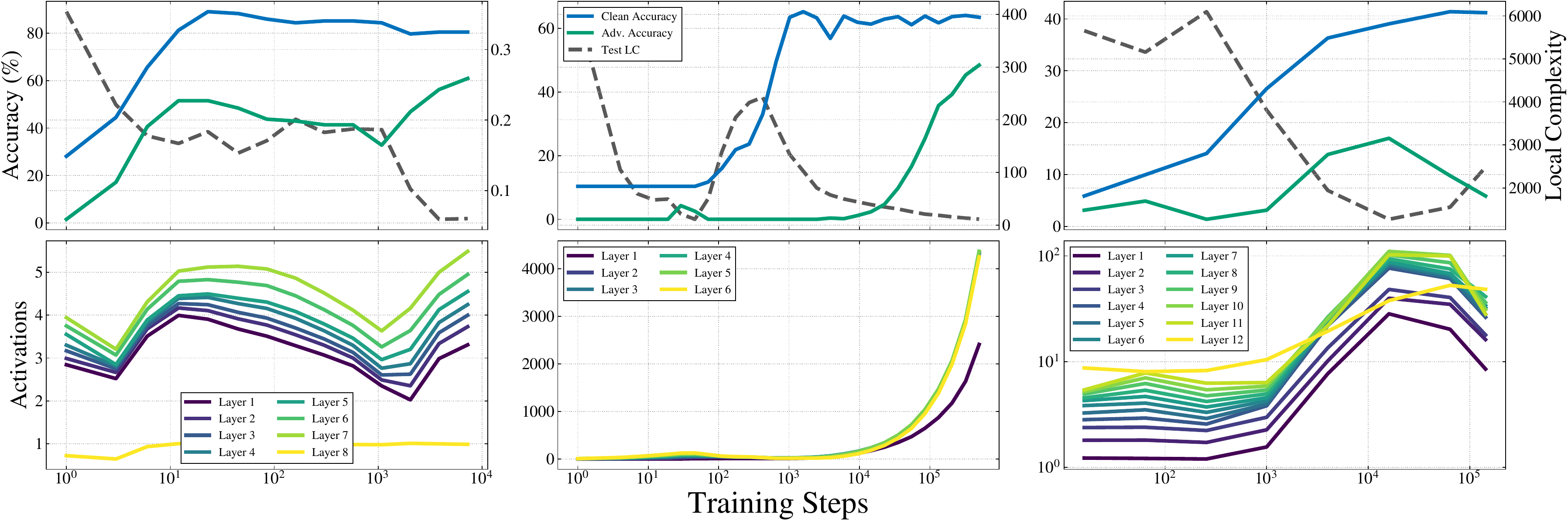}
\caption{Joint emergence of massive activations, region migration and adversarial robustness (evaluated with AutoAttack) in MLP on MNIST with $\epsilon=0.03$ (left), GPT on Shakespeare Text with $\epsilon=0.01$ (middle) Pythia 160M on The Pile with $\epsilon=0.001$ (right). The top row shows the accuracy and adversarial accuracy on the test set along with the LC (on the mirrored axis) throughout the training. The bottom row shows the sum of the three largest activations in the representation across all the layers.\vspace{-2.0em}}
\label{fig:mlp_gpt_pythia}
\end{figure}

\vspace{-1.0em}
\section{Massive Activations Induce Region Migration}
\label{sec:ma-cause-rm}
\vspace{-0.3em}

In this section, we claim that massive activations are the mechanistic cause for region migration in deep networks. The following sections detail this mechanistic connection along with extensive experimental support. Additionally, we show that this connection is not specific to any architecture or dataset and can be observed in residual MLP on MNIST, GPT on Shakespeare Text, and Pythia 160M on The Pile as depicted in Figure \ref{fig:mlp_gpt_pythia} and in ResNet18 on CIFAR10, ViT Tiny on CIFAR and ViT Small on Imagenette as illustrated in Figure \ref{fig:resnet_vit}. All the experimental details and computation of the metrics are provided in Appendix \ref{sec:exp_details}.

\subsection{Early Emergence and Persistence Through Residual Stream}

\begin{figure}[t]
\centering
\begin{minipage}{0.32\textwidth}
  \centering{\includegraphics[width=\linewidth]{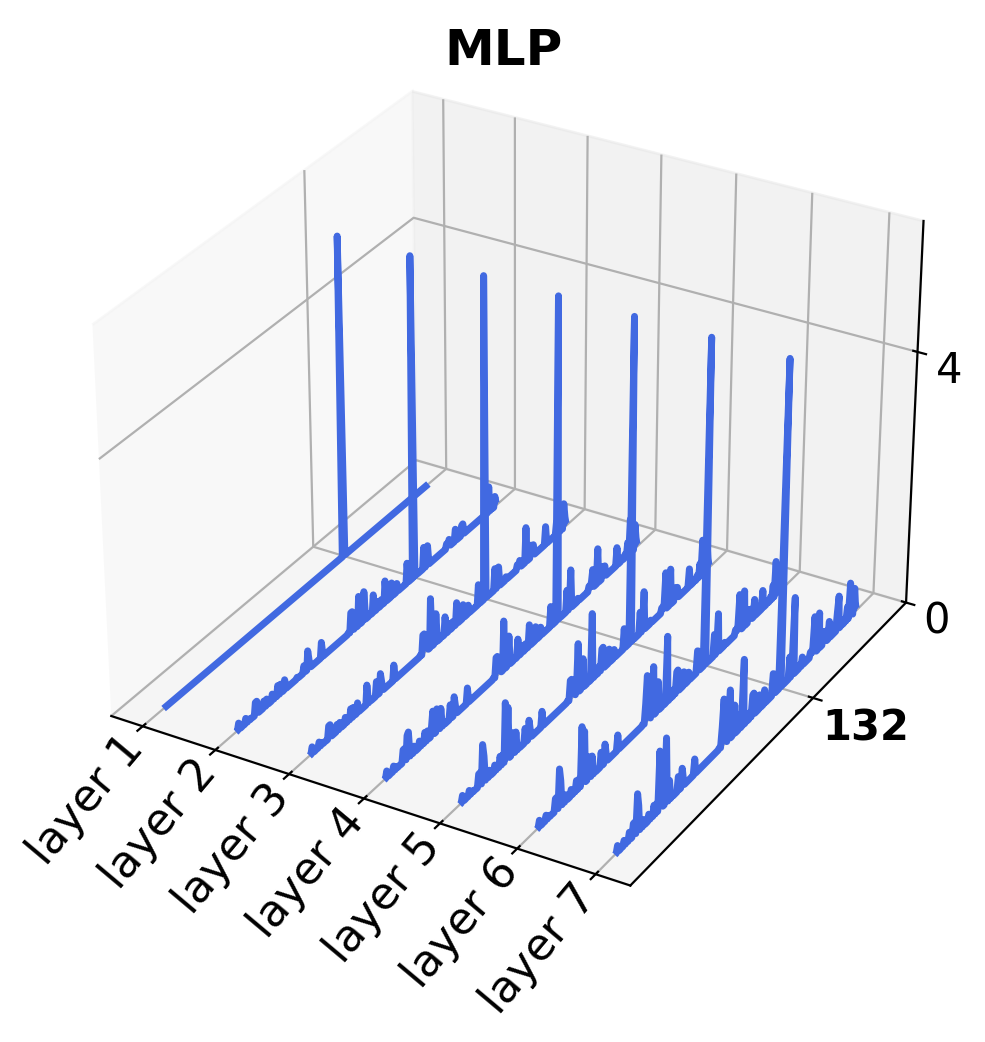}}
\end{minipage}
\begin{minipage}{0.32\textwidth}
    \centering{\includegraphics[width=\linewidth]{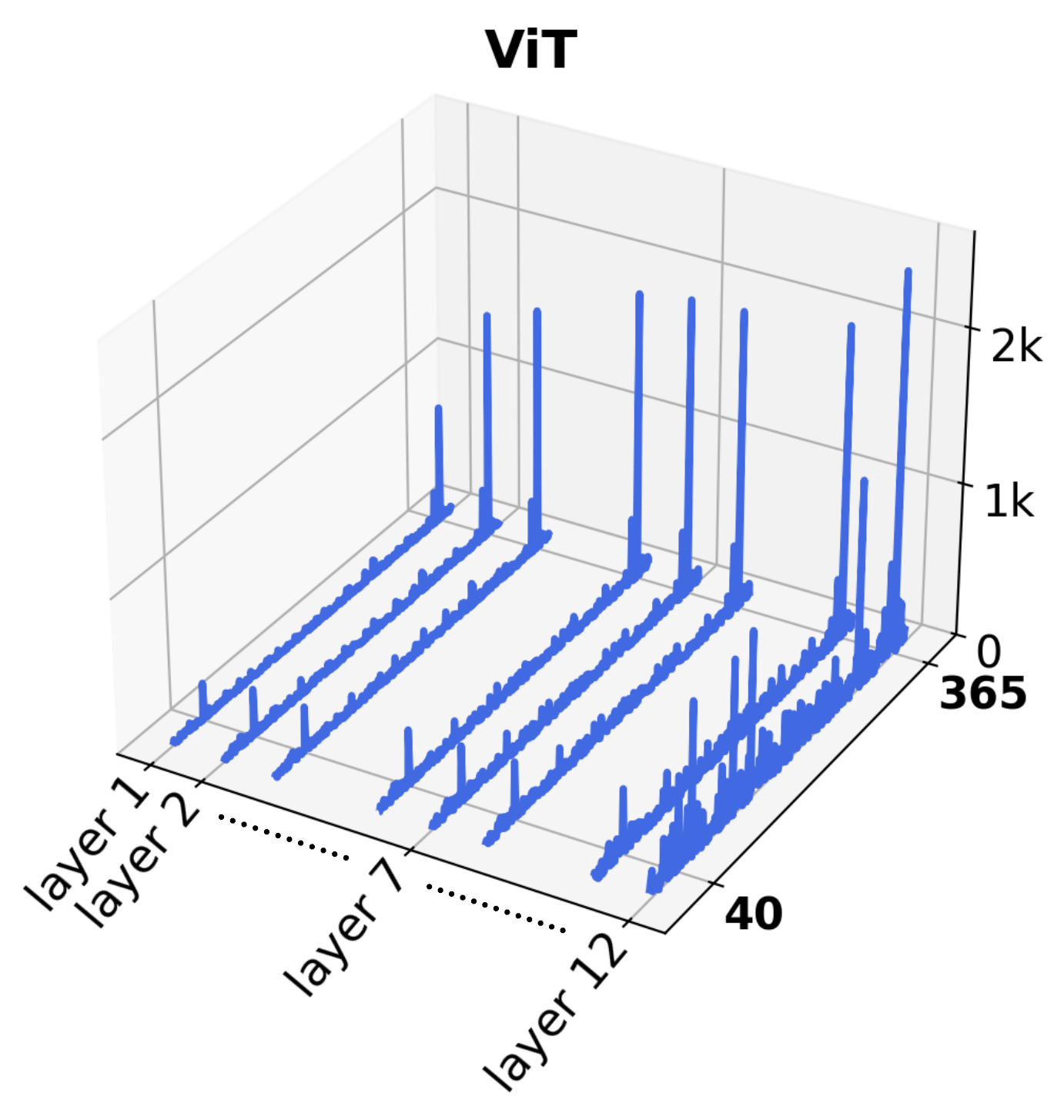}}
\end{minipage}
\begin{minipage}{0.32\textwidth}
    \centering{\includegraphics[width=\linewidth]{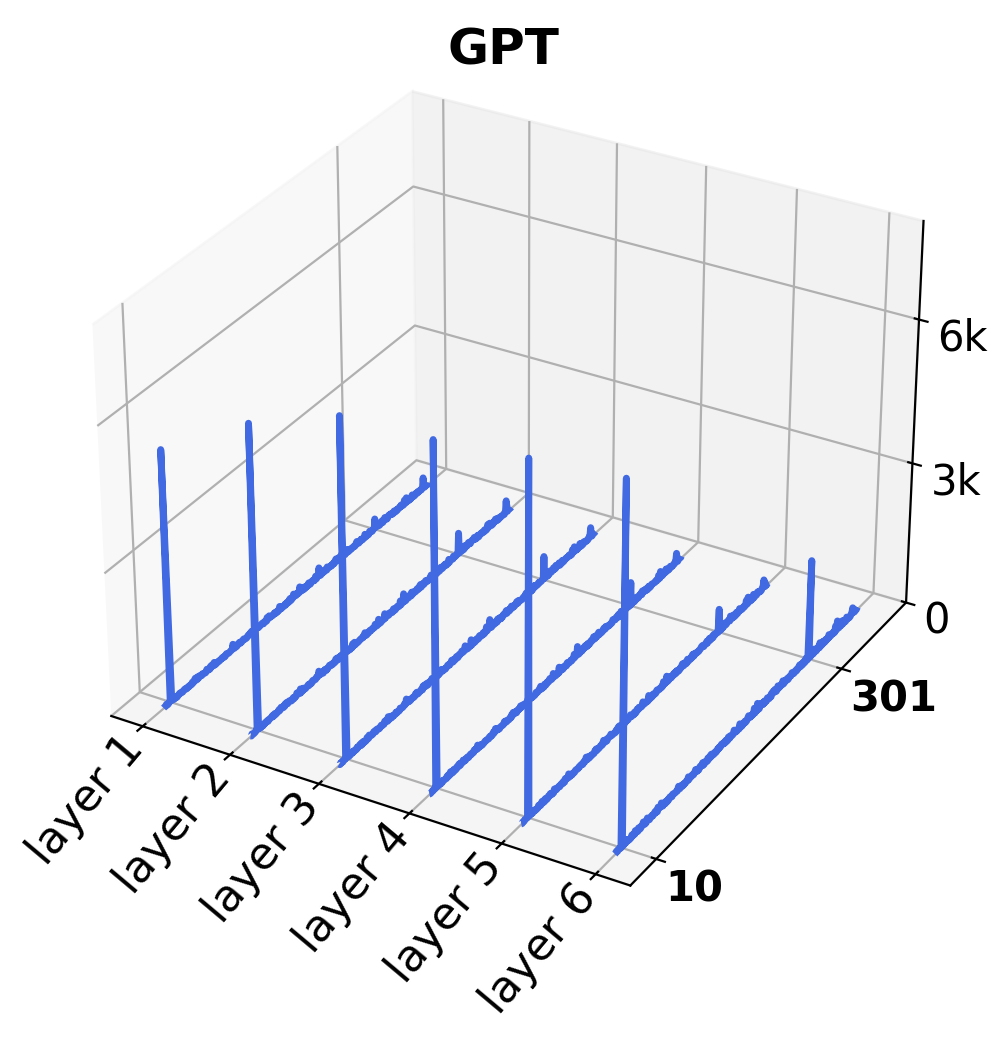}}
\end{minipage}
\vspace{-1.0em}
\caption{Visualization of massive activations in the hidden representations of residual MLP trained on MNIST (left), the CLS token representations of ViT Small trained on Imagenette (middle) and the first token representations of GPT trained on Shakespeare Text (right). Massive activations emerge as early as in the first layer and propagate through the network via the residual stream. In the case of ViT Small the massive activations amplify during the first few layers after the emergence.}
\label{fig:vit-gpt-layer-ma}
\end{figure}

Massive activations typically emerge in early layers of the network depending on the architecture. For instance in GPT (Figure \ref{fig:mlp_gpt_pythia}, middle), Pythia 160M (Figure \ref{fig:mlp_gpt_pythia}, right) and ViTs (Figure \ref{fig:resnet_vit}, middle and right) massive activations emerge in early layers. On the other hand, for ResNet18 (Figure \ref{fig:resnet_vit}, left), they do not have an exact emergence point, but gradually increase with each layer. Once they emerge, massive activations live in the representation from that point on and pass through the network via the residual stream. Figure \ref{fig:vit-gpt-layer-ma}, visualizes this mechanism via the hidden representations of residual MLP on MNIST (left), the CLS token representation of the ViT Small on Imagenette (middle) and the first token representation of GPT on Shakespeare text (right) across the network layers. In Appendix \ref{sec:ma-non-res-nets} we discuss the setting when residual connections are removed and show that massive activations emerge in early layers, however their propagation through the network largely disappears.

\subsection{Theoretical Characterization of Region Migration}

This section develops a theoretical characterization of region migration through massive activations. We first consider region migration within a single hidden layer. Let $\mW^{(\ell)} = [\vw_{1}^{(\ell)}, \dots, \vw_{D^{(\ell)}}^{(\ell)}]$ and $\vb^{(\ell)} = [b_{1}^{(\ell)}, \dots, b_{D^{(\ell)}}^{(\ell)}]$ be the parameters of layer $\ell$ and $\vx^{(\ell-1)}$ be its input representation, where $D^{(\ell)}$ is the output dimensionality of layer $\ell$. For clarity and ease of analysis we assume a ReLU activation function in the network. Given this setup, in the input space partition of layer $\ell$, the neuron $i$ induces the following hyperplane
$\mathcal{H}_i^{(\ell)}
  =
  \{
      \vz:
      {\vw_i^{(\ell)}}^\top\vz+b_i^{(\ell)}=0
  \}.
$
The LC of this layer is quantified by how many such hyperplanes lie close to the input representation $\vx^{(\ell-1)}$. This dependence can be described through the distance from the representation to each neuron hyperplane given by
\begin{equation}
  \mathrm{dist}(\vx^{(\ell-1)},\mathcal{H}_i^{(\ell)})
  =
  \frac{
      |
          {\vw_i^{(\ell)}}^\top\vx^{(\ell-1)}
          +b_i^{(\ell)}
      |
  }{\|\vw_i^{(\ell)}\|}.
  \label{eq:layer-hyperplane-distance}
\end{equation}
Therefore, LC and region migration in a single layer can be characterised as the growth of these distances. The increase of these distances, would mean that representations move away from nonlinearity boundaries and reside inside large regions.

Now suppose that the input representation $\vx^{(\ell-1)}$ has large activations. These activations would increase the distance between the representation and the neuron hyperplanes, hence displacing them into large regions and decreasing LC. To formalize this mechanism, Proposition \ref{prop:ma-layer-hyp} derives the activation magnitude required to ensure any desired distance $\delta>0$ from nonlinearity boundaries. Throughout the analysis, $\vv_{-k}$ denotes the vector obtained from $\vv$ by setting its $k$th coordinate to zero.

\begin{restatable}{proposition}{SingleLayerMA}
\label{prop:ma-layer-hyp}
    Let $0 < \eta < 1$. Assume that the massive activation at $k$th entry dominates the representation
    $$\|\vx^{(\ell - 1)}_{-k}\|<\eta|\vx^{(\ell-1)}_k|$$ and assume the neuron hyperplane parameters do not cancel out the dominance of the massive, $$|[\vw_i^{(\ell)}]_k|>\eta\|\vw^{(\ell)}_{i, -k}\|, \;\;\; i=1, \dots, D^{(\ell)}.$$
    Then $\vx^{(\ell - 1)}$ is more than $\delta>0$ distance away from  every hyperplane in the layer whenever
    \begin{equation}
        |\vx^{(\ell-1)}_k| > \max_{i=1, \dots, D^{(\ell)}}\frac{\delta\|\vw_i^{(\ell)}\|+|b_i^{(\ell)}|}{|[\vw_i^{(\ell)}]_k|- \eta\|\vw^{(\ell)}_{i, -k}\|}.
    \end{equation}
\end{restatable}
\begin{proof}
    Given in \ref{app:ma-bounds}.
\end{proof}

As massive activations emerge early in the network and travel through the residual stream, which supplies it to the inputs of the subsequent layers, they push the data points away from neuron hyperplanes in each layer, resulting in decreased LC and region migration in all layers, and hence for the whole network.

\begin{restatable}{theorem}{FullNetworkMA}
\label{thm:ma-full-net}
    Let $\alpha = \max_{\ell=1, \dots, L}\|\mW^{(\ell)}\|$ and $0 < \eta < 1$. Assume that the massive activation at entry $k$ dominates the representation at every layer
    $$
        \|\vx^{(\ell - 1)}_{-k}\|<\eta|\vx^{(\ell-1)}_k|, \;\;\; \ell = 1, \dots, L
    $$
    and every neuron hyperplane parameters do not cancel out the dominance of the massive,
    $$
        |[\vw_i^{(\ell)}]_k|>\eta\|\vw^{(\ell)}_{i, -k}\|, \;\;\; \ell = 1, \dots, L, \;\;\; i = 1, \dots, D^{(\ell)}. 
    $$
    Then the propagated ball of radius $\delta>0$ centered at $\vx^{(0)}$ does not intersect any neuron hyperplane whenever
    \begin{equation*}
        \vx^{(0)}_{k} > \max_{\ell, i}\frac{(1 + \alpha)^{(\ell - 1)}\delta\|\vw_i^{(\ell)}\|+|b_i^{(\ell)}|}{|[\vw_i^{(\ell)}]_k|- \eta\|\vw^{(\ell)}_{i, -k}\|}
    \end{equation*}
\end{restatable}
\begin{proof}
    Given in \ref{app:ma-bounds}.
\end{proof}
Under the stated assumptions, Theorem \ref{thm:ma-full-net} illustrates, that for a given distance $\delta>0$ a large enough activation magnitude would ensure that nonlinearity boundaries in the input space are further than $\delta$ from the input point. We claim that this is the mechanistic connection between massive activations and region migration. We provide additional discussion about the theoretical setup and the proofs in Appendix \ref{sec:rm-guar}.




\begin{wraptable}{r}{0.5\textwidth}
    \vspace{-2.2em}
    \centering
    \caption{Effect of zeroing dominant versus randomly selected activations on
    accuracy, adversarial consistency, and LC.
    Consistency is the percentage of predictions that remain
    unchanged under the attack, regardless of whether the clean prediction is correct.}
    \vspace{1.3em}
    \label{tab:activation-suppression-controls}
    \resizebox{0.9\linewidth}{!}{%
    \begin{tabular}{@{}llrrr@{}}
        \toprule
        Model & Condition
        & \shortstack{Accuracy ($\%$)}
        & \shortstack{Consistency ($\%$)}
        & LC \\
        \midrule
        \multirow{3}{*}{\shortstack[l]{ViT Tiny}}
            & No suppression  & 71.88 & 39.84 & 47.55 \\
            & Top 4 zeroed    & 59.38 & 21.88 & 61.97 \\
            & Random 4 zeroed & 71.72 & 38.75 & 47.20 \\
        \midrule
        \multirow{3}{*}{\shortstack[l]{GPT}}
            & No suppression  & 63.48 & 59.38 & 11.65 \\
            & Top 4 zeroed    & 25.98 &  1.95 & 43.09 \\
            & Random 4 zeroed & 63.67 & 57.30 & 11.62 \\
        \midrule
        \multirow{3}{*}{\shortstack[l]{Pythia 160M}}
            & No suppression  & 41.21 &  6.64 & 2,483.28 \\
            & Top 6 zeroed    & 21.09 &  0.98 & 3,255.83 \\
            & Random 6 zeroed & 40.55 &  6.05 & 2,517.70 \\
        \bottomrule
    \end{tabular}}
    \vspace{-1.1em}
\end{wraptable}

\subsection{Suppressing Massive Activations}
\label{sec:supp-ma}

To validate the causal link between massive activations and region migration, in this section we study the impact of suppressing massive activations on LC and prediction consistency. We found that removing some activations strongly hinders clean accuracy, making adversarial accuracy hard to interpret (since it measures the stability of correct predictions). Therefore in our suppression study we track adversarial consistency (which measures the fraction of predictions that remain unchanged under the adversarial attack) as our main metric along with accuracy and LC.

\begin{wrapfigure}{r}{0.5\textwidth}
    \centering
    \vspace{-0.0em}
    \includegraphics[width=1.0\linewidth]{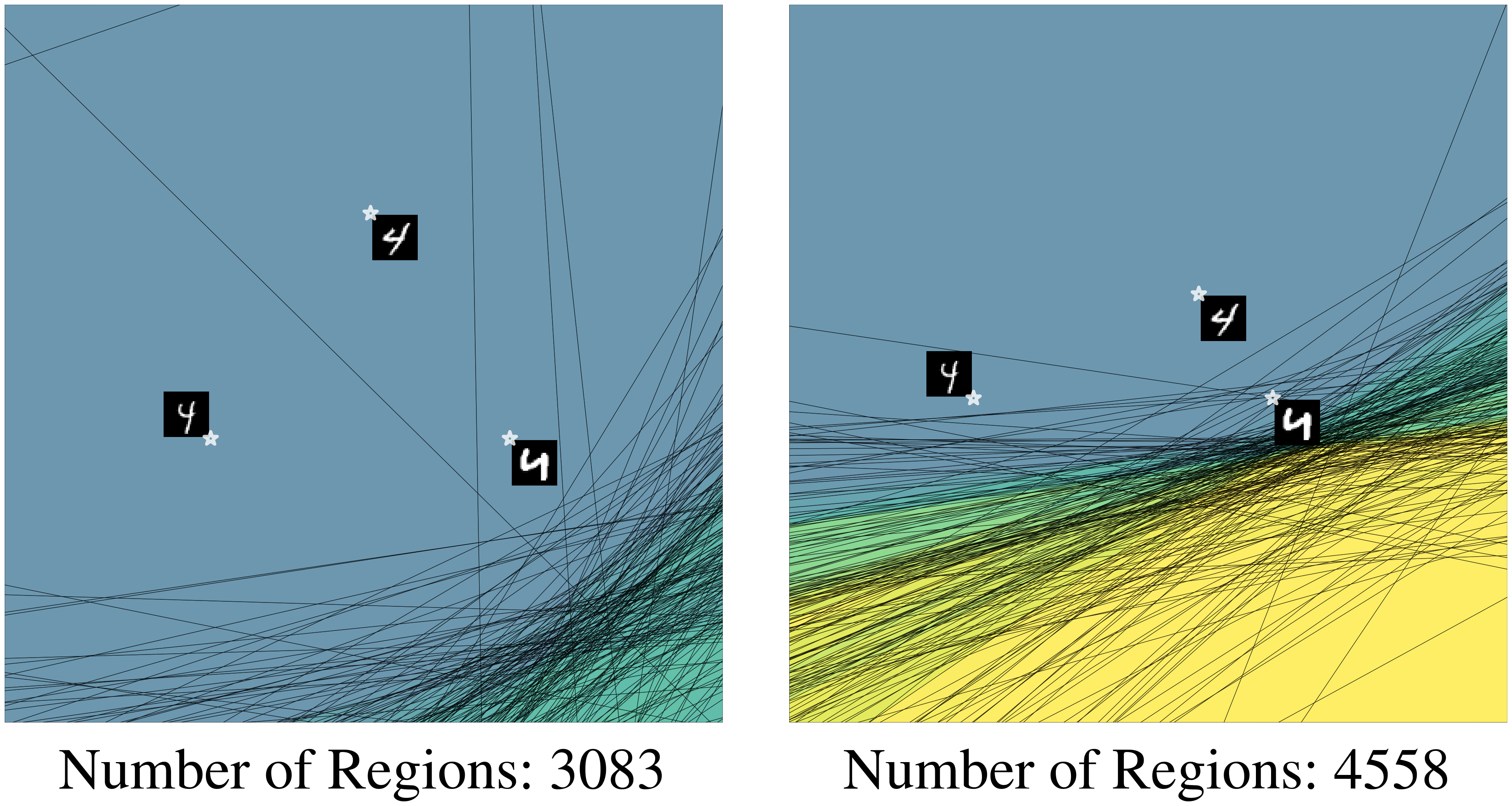}
    \caption{Removing the dominant activation from the input representation of the last layer of a $4$ depth and $200$ width residual MLP on MNIST undoes region migration. \textbf{Left:} Without any changes, serving as a reference. \textbf{Right:} The dominant activation set to zero. This Figure follows the same format as Figure \ref{fig:mlp_acc_lc_splinecam} (middle and right).}
    \label{fig:sup_ma}
    \vspace{-3.0em}
\end{wrapfigure}

Table \ref{tab:activation-suppression-controls} shows the resulting metrics when we zero out the massive activations in ViT Tiny, GPT and Pythia 160M models vs randomly selected activations. We find that consistency drops when massive activations are removed from the model, and changes only slightly when the same number of random activations are removed. This trend is accompanied by the corresponding increase in LC across all the models. These results provide further evidence for the mechanistic role of massive activations in adversarial robustness.



Additionally, Figure \ref{fig:sup_ma} provides SplineCam visualization of a trained residual MLP's last layer input space partitioning, as the dominant activations are suppressed in its input. Figure \ref{fig:sup_ma} shows that when the dominant activation is explicitly removed, region migration gets undone as depicted by the increased density of nonlinearity boundaries around data points and the number of regions in the partitioning. We provide additional examples of the input space partitioning in MLPs, when the dominant activation is decreased in Appendix \ref{sec:sup-ma-rm}.

\subsection{Which Tokens?}
\label{sec:which_tokens}

\begin{figure}[h]
\centering
\begin{minipage}{0.32\textwidth}
    \centering
    \includegraphics[width=\linewidth]{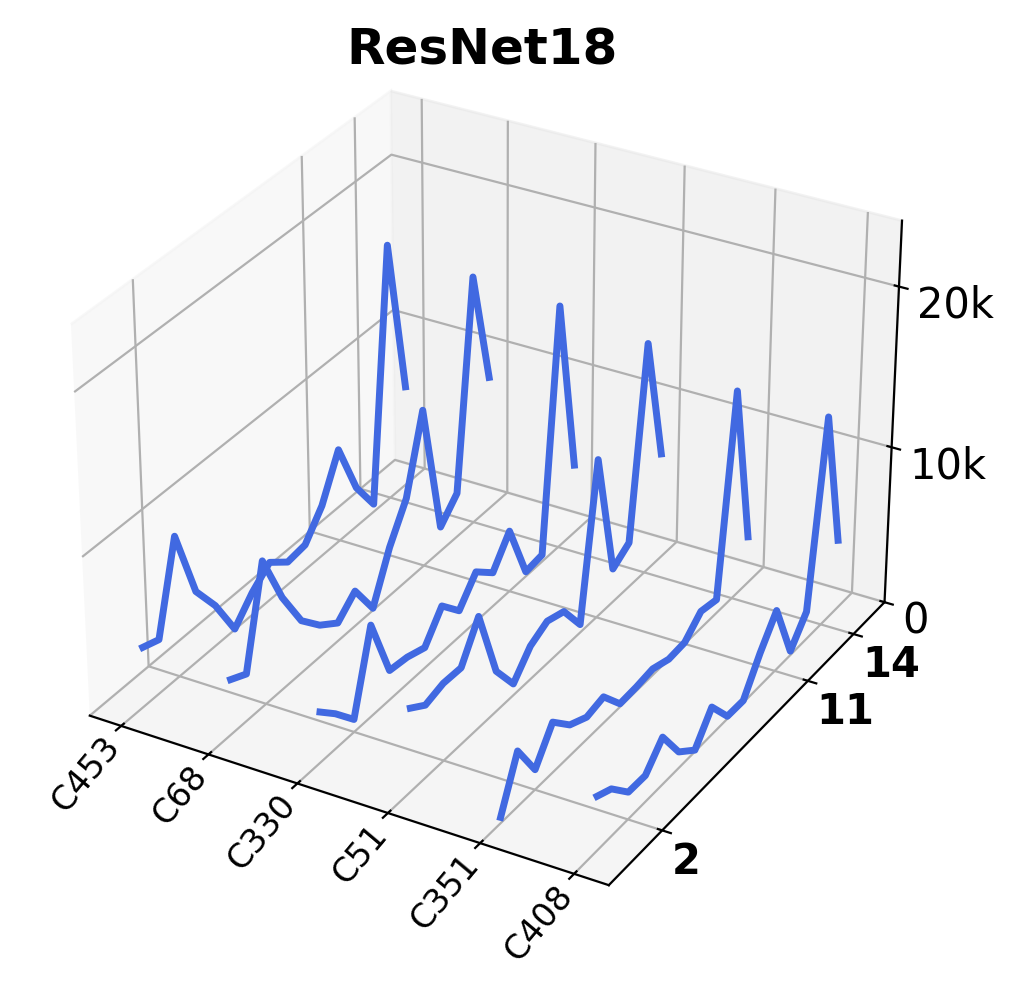}
\end{minipage}
\begin{minipage}{0.32\textwidth}
    \centering{\includegraphics[width=\linewidth]{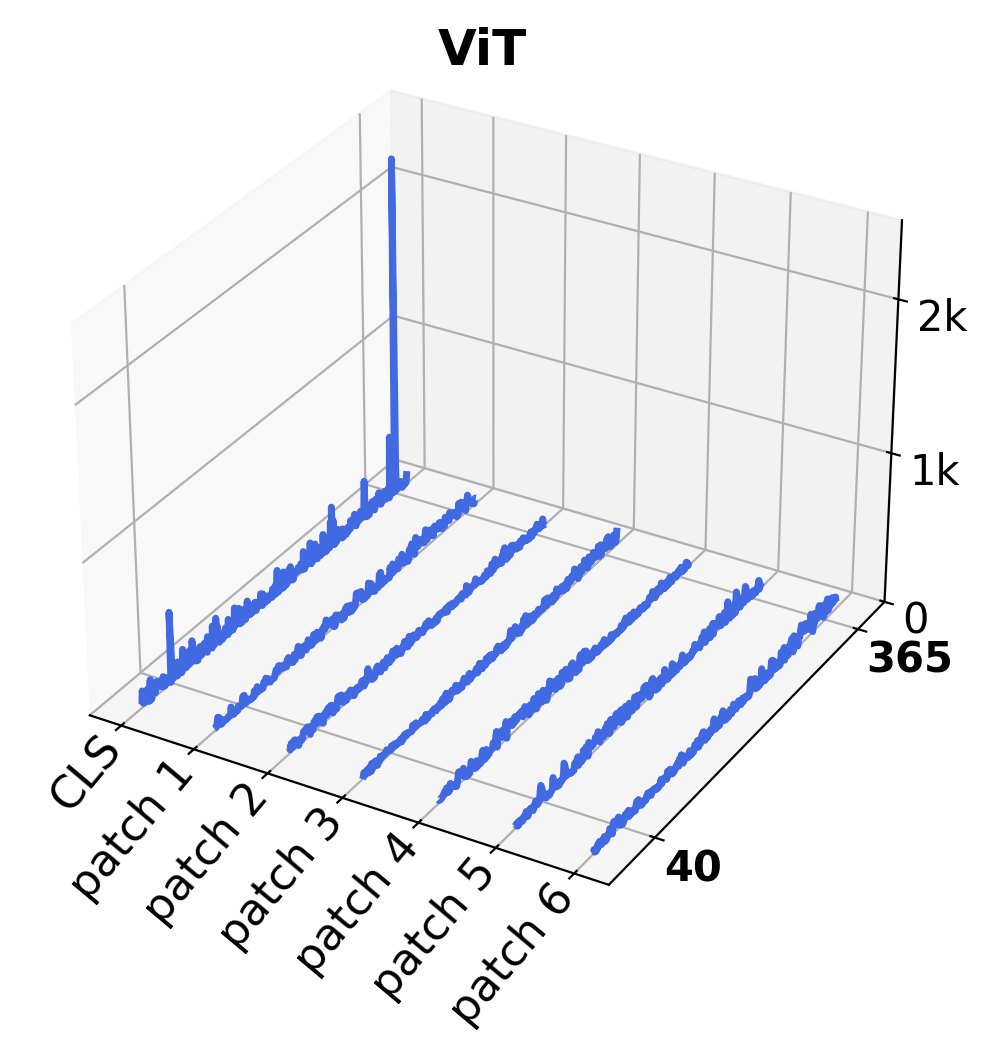}}
\end{minipage}
\begin{minipage}{0.32\textwidth}
    \centering{\includegraphics[width=\linewidth]{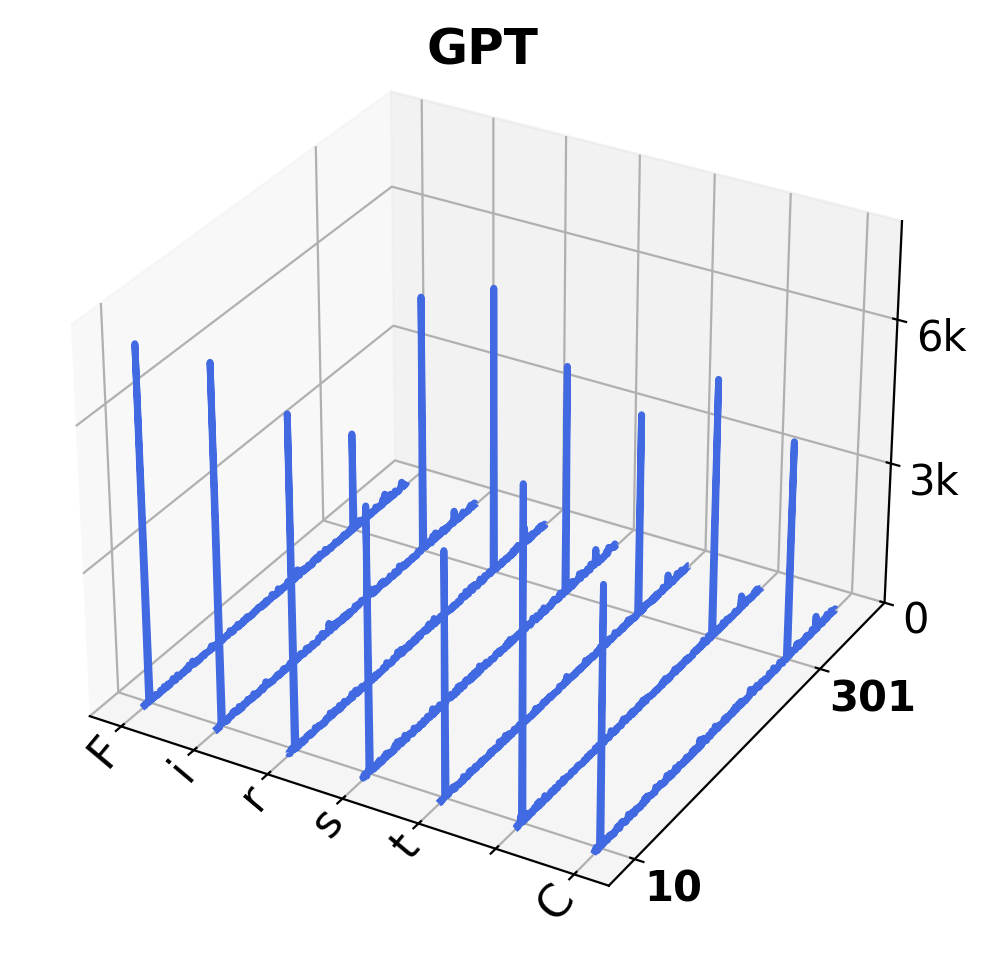}}
\end{minipage}
\caption{Visualization of the massive activations in the residual streams of ResNet18, ViT Small and GPT. \textbf{Left:} Activations of ResNet18 on CIFAR10, which contains the channels where the top $6$ (by magnitude) massive activations reside. \textbf{Middle:} Activations of ViT Small on Imagenette of the CLS token, followed by the activations of the next $6$ patch tokens. \textbf{Right:} Activations of GPT on Shakespeare Text for the first $6$ tokens in the sequence.}
\label{fig:resnet-vit-gpt-ma}
\end{figure}

Figure \ref{fig:resnet-vit-gpt-ma} shows that in ViT Small and GPT the massive activation structure is fundamentally different. In ViT Small's hidden representations they reside on the CLS token embedding, while in GPT they are present across all tokens in the hidden representation. We believe that this is not a structural coincidence, but a reflection of what region migration requires for different architectures. Recall that ViTs utilize the CLS token representation for classification, and since the MLPs (where the nonlinearities reside) in the transformer block are applied pointwise to the tokens, having the massive activation in the CLS token is sufficient for region migration. In contrast GPT would require them in all token representations because each token representation is used to make a prediction on the subsequent token. 



\begin{messagebox}
\textbf{Key Finding:} 
We illustrate that under the spline theoretic interpretation of deep learning massive activations that emerge in residual MLP, ResNet18, ViT, GPT and Pythia 160M induce region migration and co-emerge with adversarial robustness.
\end{messagebox}

\section{Region Migration and Adversarial Robustness in LLMs}
\label{sec:rm-adv-rob-llm}


In Section~\ref{sec:ma-cause-rm} we established a connection between massive activations and adversarial robustness through region migration and provided extensive experimental evidence across diverse training settings. In this section we investigate if this connection holds in the context of LLMs, containing billions of parameters. Massive activations have initially been observed in LLMs \citep{sun2024massive}, hence it is a natural question arising from our framework, whether or not the massive activations observed in LLMs are a sign of region migration and emergent robustness.

\begin{figure}[t]
\centering
\includegraphics[width=1.0\linewidth]{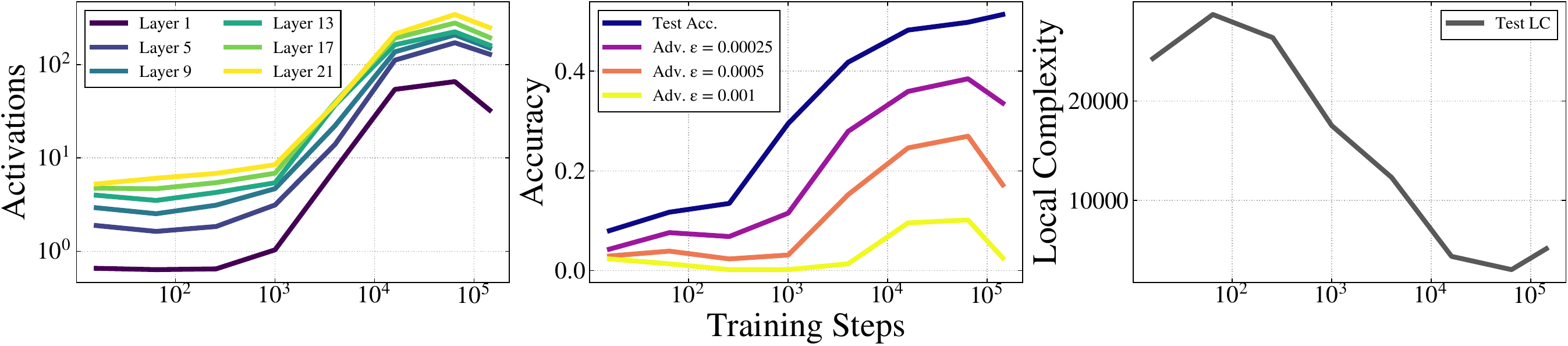}
\caption{Evolution of top activations, adversarial accuracy and LC in Pythia 1.4B. Massive activations (left) co-occur with region migration, manifested by the decreasing LC (right) and robustness to adversarial perturbations (middle). The perturbations are constructed with AutoAttack detailed in Appendix \ref{sec:metrics}.}
\label{fig:pythia_ma_advacc_lc}
\end{figure}

\subsection{Pythia Suite}

We study the Pythia suite of LLMs \citep{biderman2023pythia}, a family of decoder only transformers pretrained on The Pile~\citep{biderman2023pythia} dataset. We use the checkpoints available throughout pretraining to monitor the emergence of massive activations and their relation to LC and adversarial robustness in the same dataset.

\paragraph{Experiment Setup.} Similar to the prior experimental setup the sum of the largest three activations in the hidden token representation is recorded and averaged across tokens and samples across all layers in the network. The adversarial perturbations are applied to token embeddings and adversarial accuracy is measured on the next-token prediction task, similar to the GPT on Shakespeare Text setting. The attack configuration is detailed in Appendix \ref{sec:metrics}.

\paragraph{Results.} Figure \ref{fig:pythia_ma_advacc_lc} presents the results of this experiment on Pythia $1.4$B and illustrates that the emergence of massive activations (left) closely tracks the descent of LC (right) and robustness to adversarial perturbation (middle). We observe the same trend across different model sizes in the Pythia suite. These results demonstrate that massive activations, originally observed in LLMs, are indeed connected to the emergence of adversarial robustness. In Figure \ref{fig:pythia-suite-ma}, we provide the massive activations in the residual stream of the Pythia models across different scales. Notably, all token representations possess massive activations, which is consistent with the discussion for GPT on Shakespeare Text setting detailed in Section~\ref{sec:which_tokens}. Appendix \ref{sec:ma-qwen-gemma} provides additional evidence for the emergence of per-token massive activations for Qwen 2.5 \citep{qwen2024qwen2} and Gemma 3 \citep{Kamath2025Gemma3T} pretrained model families.

\subsection{Why Is Delayed Robustness Less Evident in LLMs?}

In Figure \ref{fig:pythia_ma_advacc_lc}, we observe that region migration occurs relatively earlier during training, compared to the experiments in Section~\ref{sec:ma-cause-rm}, where adversarial accuracy starts increasing long after the train and test accuracies plateau. We hypothesise that this is due to the fact that the dataset size of the Pythia suite is significantly larger than in any of the training settings in Section~\ref{sec:ma-cause-rm} and it has been observed in \citet{humayun2023splinecam} that this factor accelerates the occurrence of delayed robustness.

\begin{messagebox}
  \textbf{Key Finding:} We find that massive activations in LLMs actually induce region migration and emerge alongside robustness to perturbations of the input sequence embeddings during pretraining.
\end{messagebox}

\section{Discussion and Future Work}

In this work, we have uncovered a connection between massive activations and adversarial robustness in deep learning. We have shown that massive activations push the representations of the network away from the neuron hyperplanes, which define the input space partitioning boundaries, inducing region migration, which is connected to adversarial robustness. We demonstrated that this connection is not specific to any particular architecture or dataset configuration and can be observed in residual MLPs, ResNet18, ViTs, GPTs, and can even be extended to LLMs pretrained on diverse text data.

\begin{figure}[t]
  \begin{minipage}{0.24\textwidth}
      \centering{\includegraphics[width=\linewidth]{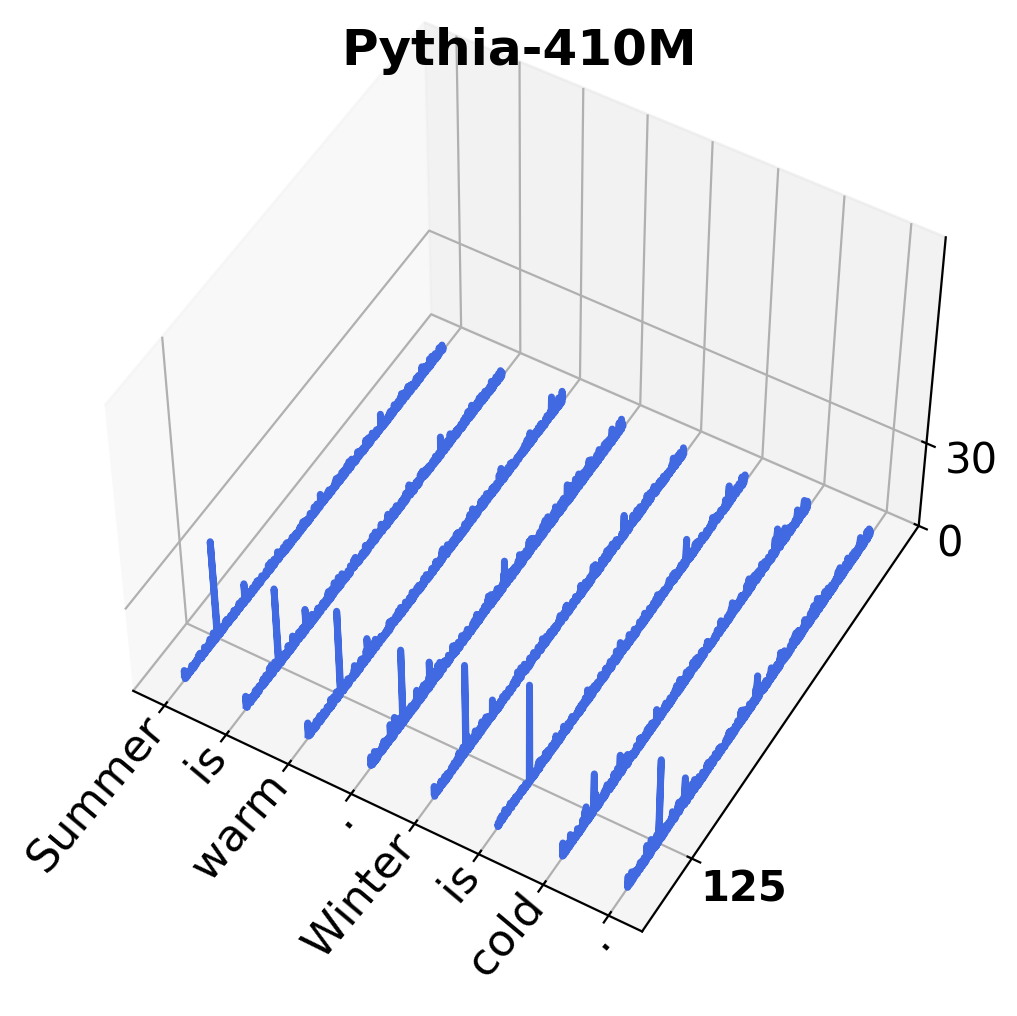}}
  \end{minipage}
  \begin{minipage}{0.24\textwidth}
      \centering{\includegraphics[width=\linewidth]{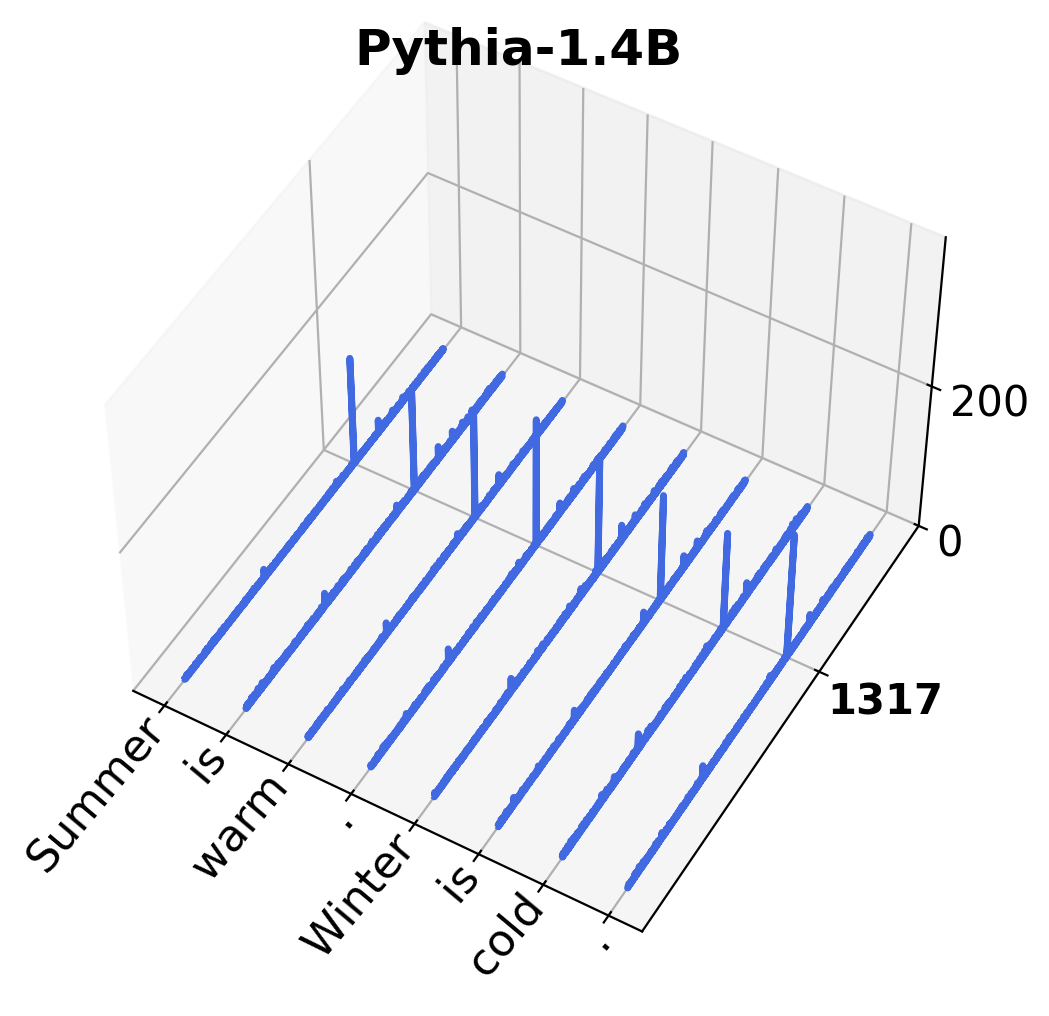}}
  \end{minipage}
  \begin{minipage}{0.25\textwidth}
      \centering{\includegraphics[width=\linewidth]{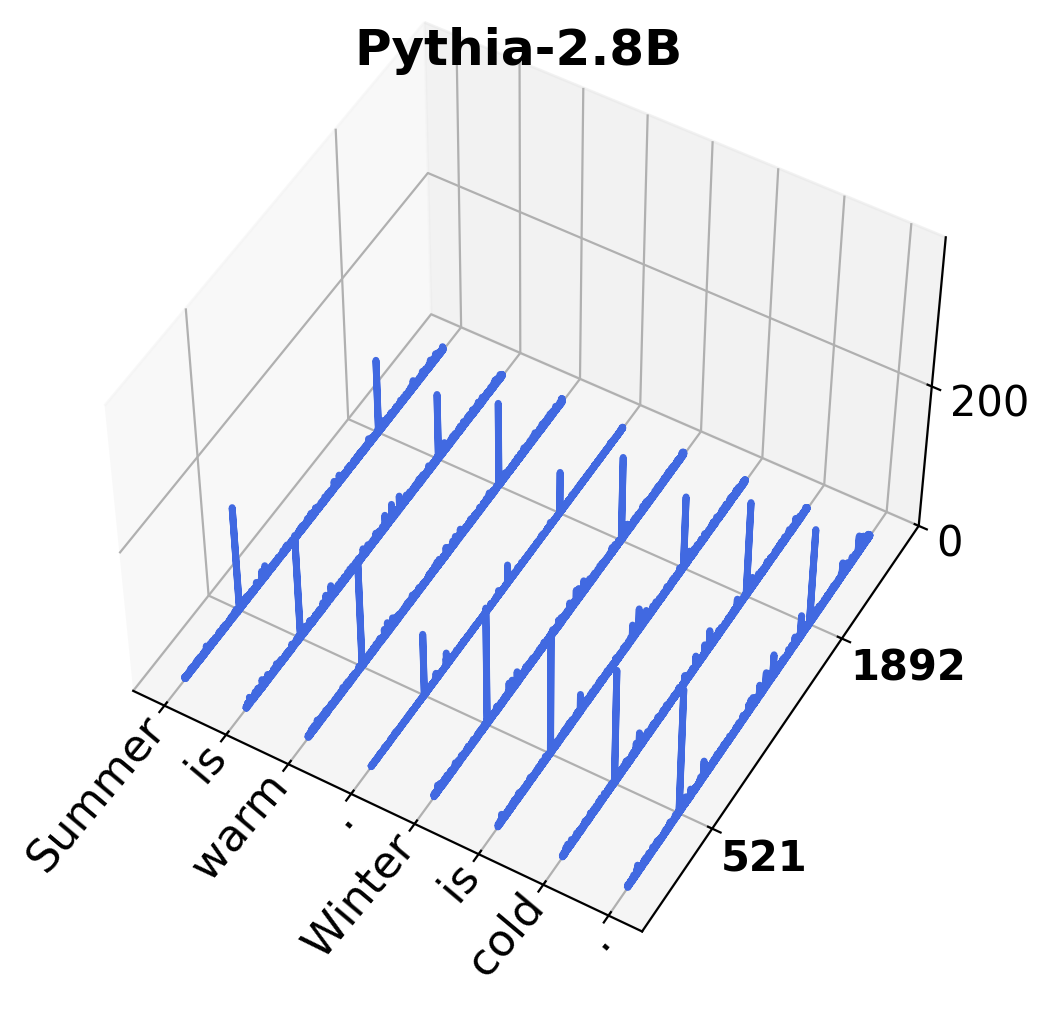}}
  \end{minipage}
  \begin{minipage}{0.24\textwidth}
      \centering{\includegraphics[width=\linewidth]{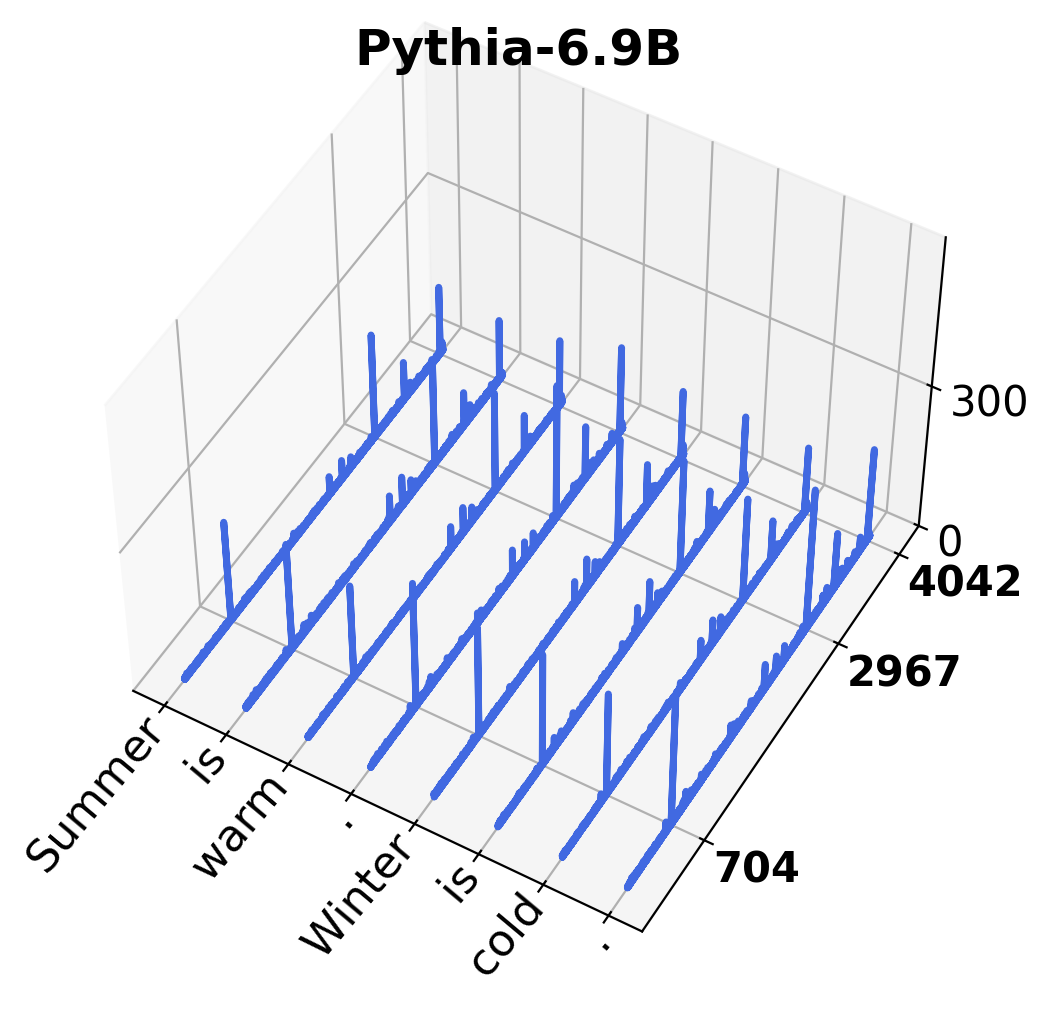}}
  \end{minipage}
  \caption{The activation magnitudes in the hidden representations of the Pythia suite across different model sizes.}
  \label{fig:pythia-suite-ma}
\end{figure}

\paragraph{Are Massive Activations Useful?} The existing literature offers no concrete explanations for the role of massive activations in deep learning and whether they are useful or undesirable. \citet{sun2024massive} showed that they act as fixed biases in the model and their removal results in significant performance degradation. Massive activations also induce attention sinks \citep{xiao2024efficient}, which have been argued to prevent over-mixing in the attention operation \citep{barbero2025why}. On the other hand, \citet{xiao2023smoothquant, dettmers2022gptint} show that large magnitude activations make quantising neural networks more challenging. Our work adds a new perspective to this discussion, suggesting that massive activations act as a mechanism to simplify the local input-output mapping of neural networks, as quantified by the LC progress measure, which empirically accompanies increased robustness.

\paragraph{Delayed Robustness in LLMs.} We found that delayed robustness occurs in LLM pretraining, with the rise of massive activations. This finding shows that delayed robustness holds beyond the standard training settings, where it was originally discovered \citep{humayun2024deep}, expanding its applicability to large-scale deep learning.

\paragraph{Massive Activations in LLMs.} In Section~\ref{sec:rm-adv-rob-llm} we showed that massive activations emerging in the Pythia suite induce region migration. Our framework suggests that per-token massive activations are associated with robust next token prediction, illustrated in Pythia family. However, we highlight that \citet{sun2024massive} found that not all model families have a per-token massive activation structure (i.e. LLaMA2 \citep{touvron2023llama}). We believe that follow-up research could investigate if these LLM families experience delayed robustness and their vulnerability to adversarial attacks. Additionally, we believe an interesting direction of future study is to develop pretraining prescriptions that better exploit delayed robustness and improve adversarial robustness.


\section{Limitations}

While this work sheds new light on the role of massive activations and why they are useful in deep learning, we identify the following limitations. It is not clear what drives the emergence of massive activations in the training and why deep networks construct a small number of exceptionally large activations, rather than distributing it across many moderately large activations. The theoretical setup analysing the connection between massive activations and region migration is based on a fully connected neural network with residual connections and does not cover operations like attention and normalization layers, which are commonplace in deep learning.

\section{Conclusion}

In this paper, we showed why massive activations~\citep{sun2024massive} are useful in deep learning, via connecting them to delayed robustness~\citep{humayun2024deep}. Under the spline theoretic view of neural networks, we showed that massive activations reduce the network's LC, by displacing representations away from the network's nonlinearity boundaries, inducing region migration, which coincides with emergent robustness. We supported this claim through extensive experimental evidence and derived theoretical arguments that guarantee the simplification of the input space partitioning around data points if the activations in the representations are sufficiently large. We further showed that our findings are also relevant in the context of LLMs, by uncovering that the emergence of massive activations in LLMs closely tracks region migration and robustness to perturbations of the input token embeddings.



\bibliography{thesis}
\bibliographystyle{iclr2027_conference}

\appendix

\section{Review of Spline Theory of Deep Networks}
\label{sec:spline_th}

This section provides the preliminaries of the Spline Theory of Deep Networks~\citep{balestriero2018spline, balestriero2020mad}, which builds a mathematical bridge between neural networks and splines, offering a powerful gateway to analyse their inner workings \citep{balestriero2025geometry, balestriero2022batch, humayun2024deep}. Section~\ref{sec:splines} introduces splines, with the emphasis on the max affine splines, which will serve as the building blocks to establish the connection to neural networks. Section~\ref{sec:dn-maso} presents the precise connection between the composition of max affine spline operators and deep networks and Section~\ref{sec:inp-space-geo} presents the geometric properties of deep networks composed of splines.

\subsection{Splines}
\label{sec:splines}

This section presents the terminology and definitions for splines. Specifically, we will discuss affine splines (also called linear or first order splines). The following discussion defines affine splines and max affine splines in the multivariate input-output setting.

\subsubsection{Affine Splines}

We start by defining multivariate \textit{affine spline functions}.
\begin{definition}[Multivariate Affine Spline Functions]
    Let $\Omega = \{\omega_1, \dots, \omega_R\}$ be a partition of a domain $\mathbb{R}^{D}$ and let $\Phi=\{\phi_{1}, \dots, \phi_{R}\}$ be mappings from $\mathbb{R}^{D}$ to $\mathbb{R}$ for each partition $\omega_{r}$ via $\phi_{r}(\vx) = \alpha_{r}^\top\vx + \beta_{r}$, where $\vx \in \omega_{r}$ and $\alpha \in \mathbb{R}^{R \times D}$ and $\beta \in \mathbb{R}^{R}$ are the parameters of the affine mappings. Under this setup an affine spline function is defined by
    \begin{equation}
        s[\alpha, \beta, \Omega](\vx) = \sum_{r=1}^{R}(\alpha_r^\top \vx + \beta_r)\mathbf{1}(\vx \in \omega_r)
    \end{equation}
    where $\mathbf{1}(\vx \in \omega_r)$ is the indicator function.
\end{definition}
When approximating functions with splines, a substantial challenge is the need to jointly optimize for the partitioning of the input space $\mathbb{R}^{D}$ and for the affine mapping parameters $\alpha$ and $\beta$ \citep{balestriero2018spline}. Next section discusses how to resolve the issue of joint optimization, by enforcing convexity on the affine spline functions.

\subsubsection{Max Affine Splines}

A simplification of the joint optimization arises when affine spline functions are constrained to be convex, in which case the function can always be written as \textit{max affine spline} \citep{balestriero2018spline, hannah2013multivariate, magnani2009convex}.

\begin{definition}[Max affine spline functions]
    Let $\alpha \in \mathbb{R}^{R \times D}$ be a matrix of hyperplane "slopes" and $\beta \in \mathbb{R}^R$ be a vector of "offsets", then a max affine spline function is defined as
    \begin{equation}
        s_{\alpha, \beta, \Omega}(\vx) = \max_{r=1, \cdots, R}{\alpha_{r}^\top\vx + \beta_r}
    \end{equation}
\end{definition}

A useful property of max affine spline functions is that they are completely determined by their affine mapping parameters $\alpha$ and $\beta$, and the selection of the partition $\Omega$ is automatically determined by the choices of $\alpha$ and $\beta$ \citep{balestriero2018spline}. Hence a max affine spline function can be denoted as $s_{\alpha, \beta}$, as $\Omega$ is inferred by $\alpha$ and $\beta$. This provides a resolution for the joint optimization of the partition and the affine mapping parameters discussed in the previous section.

Next we present \textit{max affine spline operators} (MASOs), which are a natural extension of max affine splines into the multivariate outputs, by concatenating $K$ independent max affine spline functions.

\begin{definition}[Max Affine Spline Operators]
    Let $A \in \mathbb{R}^{K \times R \times D}$ and $B \in \mathbb{R}^{K \times R}$ be the parameters, then max affine spline operator is defined as
    \begin{equation}
        S_{A, B}(\vx) =
        \begin{bmatrix}
            \max_{r=1, \cdots, R}{A_{1,r}^\top\vx + B_{1,r}} \\
            \vdots \\
            \max_{r=1, \cdots, R}{A_{K,r}^\top\vx + B_{K,r}}
        \end{bmatrix}
    \end{equation}
\end{definition}

A MASO is always a piecewise affine and convex with respect to each output dimension \citep{balestriero2020mad}.

\subsection{Deep Networks as Compositions of MASOs}
\label{sec:dn-maso}

In this section we state the connection between neural networks and spline theory. The central result in \citet{balestriero2018spline, balestriero2020mad} is that a broad range of layers used in neural networks can be represented as MASOs, and hence a deep neural network is a composition of MASOs.

\subsubsection{Fully Connected Layers and Activation Functions}

\begin{proposition}[Fully Connected/Convolutional Layers]
\label{prop:full-layer-maso}
  Any fully connected or convolutional layer is an affine mapping and hence a MASO with $R=1$.
\end{proposition}

A fully connected layer with parameters $\mW$ and $\vb$ is represented by $\mW^\top\vx+\vb$, which is a single affine mapping, that is it is equivalent to $S_{\mW, \vb}(\vx)$ with $R=1$. This reasoning naturally extends to the convolutional layer.

\begin{proposition}[Activation Functions]
    \label{prop:act_maso}
    Any piecewise affine and convex activation function (such as \text{ReLU}, \text{Leaky ReLU} or \text{Absolute Value}) is a MASO.
\end{proposition}

To exemplify the intuition behind this result, consider the ReLU activation $\text{ReLU}(\vx) = \max\{\vx, 0\}$, that is applied elementwise to $\vx$. The output at $k$th position is $\max\{\mathbf{e}_k^{\top}\vx, 0\}$, which is a max affine spline with $R=2$, with $A_{k,1}=0$, $A_{k,2}=\mathbf{e}^{k}$ and $B_{k,1}=B_{k,2}=0$, where $\mathbf{e}_{k}$ is the $k$th canonical basis vector. The mapping for Leaky ReLU follows the same reasoning, except $A_{k,1}=v\mathbf{e}_{k}$, where $v$ is the negative slope of the Leaky ReLU activation function. For absolute value $A_{k,1} = -\mathbf{e}_{k}$ and $A_{k,2} = \mathbf{e}_{k}$.

\subsubsection{Deep Networks}

This section presents the main result of \citet{balestriero2020mad} connecting deep networks to splines.

\begin{proposition}
\label{prop:layer-maso}
    A deep network layer constructed from any composition of fully connected or convolutional layers followed by a single activation operator is a MASO.
\end{proposition}

Proposition \ref{prop:layer-maso} shows that a deep network layer composed of a fully connected or convolutional layer followed by an activation function is a MASO. Subsequently the following theorem establishes that a deep network is a composition of MASOs.

\begin{theorem}
\label{tm:dn-maso}
    A deep neural network composed of fully connected/convolutional layers, activation functions (defined according to \ref{prop:act_maso}) is a composition of MASOs.
\end{theorem}

We note that while a layer of a network is a MASO, as illustrated by Propositions~\ref{prop:full-layer-maso}, \ref{prop:act_maso} and \ref{prop:layer-maso}, a deep network, that is, a composition of MASOs, is not a MASO in general and this is what introduces the non-convexity and the expressive power of neural networks \citep{balestriero2018spline, balestriero2020mad}. We do not provide the proofs for these theoretical results in this project for brevity and refer the reader to \citet{balestriero2020mad} for the detailed derivations.



\paragraph{General Activation Functions.} While the central building block in the Spline Theory is the MASO, which covers activation functions like ReLU, LeakyReLU and the absolute value, it does not naturally extend to smooth activation functions like sigmoid. However, since any continuous function can be approximated with arbitrary precision via affine splines, the spline theoretic view of deep networks would approximately hold \citep{balestriero2018spline}.

\paragraph{Extension to General Neural Networks.} In addition to the discussed connection between splines and neural networks, \citet{balestriero2020mad} shows that various operations that are commonplace in deep learning, such as residual connections and pooling layers, can also be expressed as splines. On top of that \citet{balestriero2020mad} presents the detailed derivations for the neural network to spline mapping for convolutional neural networks (CNNs) \citep{lecun1998mnist}, ResNets \citep{he2016deep}, and recurrent neural networks (RNNs) \citep{graves2013generating}, illustrating that spline theory perspective is applicable in a wide range of architectures.

\subsection{Input Space Geometry}
\label{sec:inp-space-geo}

A direct consequence of the max affine spline structure of deep networks is its specific geometric structure. As discussed in previous sections, each MASO layer partitions its input space and applies a region dependent affine mapping on it. \citet{balestriero2019geometry} provides a comprehensive analysis about the geometry of the partition, revealing that it has a polytopal structure.

\begin{theorem}
\label{tm:maso-part}
    Let $f$ be a MASO layer, then it partitions its input space into a finite collection of convex polytopes.
\end{theorem}

Theorem \ref{tm:maso-part} shows that a MASO layer or a network layer divides its input space into convex polytopal regions. A composition of $L$ MASO layers produces an increasingly subdivided partition of the input space, where each next layers further divides the partition induced by the previous layers \citep{balestriero2018spline}. Next we provide the result about the geometry of the input space partitioning of a deep network.

\begin{figure}[t]
    \centering
    \begin{minipage}{0.30\textwidth}
        \centering
        \includegraphics[width=\linewidth]{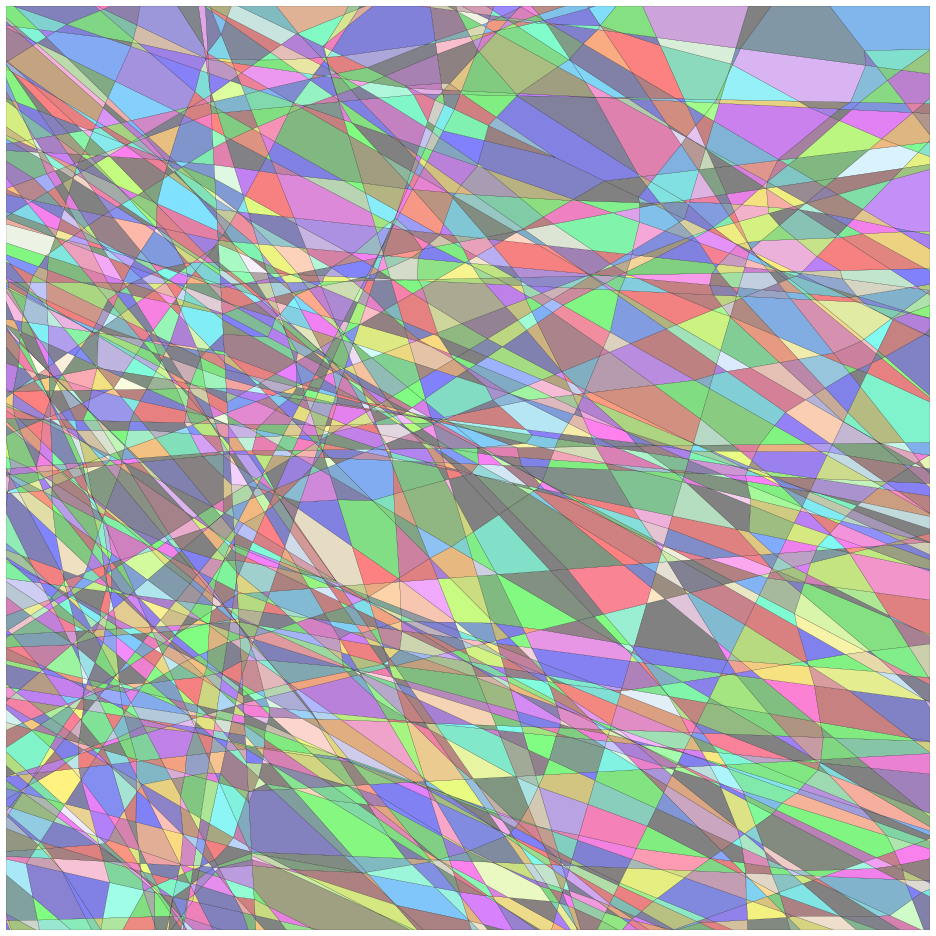}
    \end{minipage}
    \caption{Output of a SplineCam visualization. The input space partitioning on a $2$D slice of the input space from a $4$ layer ReLU MLP. On each region the network is an affine mapping. Colors are randomly selected.}
    \label{fig:part-spline-vis}
\end{figure}

\begin{theorem}[Deep Network Input Space Partitioning]
\label{tm:dn-part}
    Let $f = f^{(L)} \circ \cdots \circ f^{(1)}$ be a composition of $L$ MASO layers, then the input space partitioning consists of finite convex polytopes given by $\{\omega_r\}_{r}$ and moreover on each cell $\omega_{r}$ the deep network is an affine mapping
    \begin{equation}
        f(\vx) = \sum^{R}_{r=1} (A_{\omega_{r}}^{\top}\vx+B_{\omega_{r}})\mathbf{1}(\vx \in \omega_r)
    \end{equation}
\end{theorem}
Theorem \ref{tm:dn-part} states that an affine spline network partitions its input space into a finite number of convex polytopal regions, where within each polytopal region, the network is an affine map. That is let $\vx$ be in a partition $\omega_\vx$, then the network input-output map reduces to
\begin{equation}
    f(\vx) = A_{\omega_\vx}^\top\vx + B_{\omega_\vx}
\end{equation}
where $A_{\omega_\vx}$ and $B_{\omega_\vx}$ are the affine mapping parameters for partition $\omega_\vx$. Figure \ref{fig:part-spline-vis} provides a visualization of the input space polytopal partitioning on a toy $4$ layer ReLU multilayer perceptron. We do not provide the proofs of Theorems \ref{tm:dn-part} and \ref{tm:dn-maso}, as they have extensive derivations and refer the reader to \citet{balestriero2020mad} for the details.



\subsection{SplineCam}
\label{app:splinecam}

The spline theoretic framing of deep networks provides insights into the geometry of the input space partition, however computing and visualizing these partitions is not trivial. SplineCam \citep{humayun2023splinecam} was introduced to solve this problem, which is an exact and sampling free method for computing the geometry of deep network input space partitioning.

SplineCam is applicable for any affine spline network (as defined in Section~\ref{sec:dn-maso}). Since the neural network input space is high dimensional and cannot be easily visualized, \citet{humayun2023splinecam} propose to visualize the partition on a two dimensional slice of the input space. The output of a SplineCam visualization is a $2$D image, where each contiguous line represents a neuron hyperplane boundary and each region enclosed by the lines is a partition region where the network input-output map is an affine transformation. An example of a SplineCam visualization of a $4$ layer ReLU MLP is provided in Figure \ref{fig:part-spline-vis}.

\section{Region Migration Guarantees}
\label{sec:rm-guar}

Section~\ref{sec:ma-cause-rm} provided a mechanistic description with extensive experimental support of how massive activations emerge, propagate in the network and cause region migration. In this section we develop a theoretical understanding of region migration induced by massive activations to further support our empirical claims. In particular, we show that, large enough massive activations in the input representations of the network layer guarantee low LC.

\subsection{Setup and Assumptions}

The aim of our analysis is to study the behaviour of the region migration in deep networks under massive activations. We study a fully connected neural network with residual connections with ReLU activation function (which is consistent with the definition in Proposition~\ref{prop:act_maso}) for ease of analysis. Specifically, we consider the following network
\begin{equation}
    \vx^{(\ell)} = \text{ReLU}\left(\mW^{(\ell)\top}\vx^{(\ell - 1)} + \vb^{(\ell)}\right) + \vx^{(\ell - 1)}, \;\;\; \ell = 1, \dots, L
\end{equation}
where $\mW^{(\ell)} = [\vw_{1}^{(\ell)}, \dots, \vw_{D^{(\ell)}}^{(\ell)}]$ and $D^{(\ell)}$ is the width of layer $\ell$.

\subsubsection*{Quantifying Region Migration}

To measure region migration and LC, we quantify the distance of the representations from the network's nonlinearities. For a ReLU activation function, the nonlinearity occurs at the point where the pre-activations cross zero. In particular for a neuron $i$ in a layer $\ell$ the nonlinearity occurs when $\vw_{i}^{(\ell)\top}\vx^{(\ell - 1)} + \vb^{(\ell)}_i = 0$, hence we define the neuron hyperplane
\begin{equation}
    \mathcal{H}^{(\ell)}_{i}=\left\{\mathbf{z} : \vw_{i}^{(\ell)\top} \mathbf{z} + \vb^{(\ell)}_i = 0\right\}
\end{equation}
As we are interested in computing distances of the representations from such hyperplanes to assess how close the representation is from changing the pre-activation sign and falling into another affine mapping, Remark~\ref{rem:hyp-dist} highlights the formula of the distance of a point from a hyperplane.

\begin{remark}
\label{rem:hyp-dist}
    Let $\vw^\top\vx + b = 0$, be a hyperplane, then the distance of a point $\vx_0$ from the hyperplane is given by $\frac{|\vw^\top\vx_0+b|}{\|\vw\|}$.
\end{remark}

Hence, the distance of a representation $\vx^{(\ell)}$ at layer $\ell$ from the nonlinearity boundary $\mathcal{H}^{(\ell)}_{i}$ for neuron $i$ is given by
\begin{equation}
\label{eq:rep-dist}
    \text{dist}(\vx^{(\ell - 1)}, \mathcal{H}^{(\ell)}_{i}) = \frac{|\vw_i^{(\ell)\top}\vx^{(\ell - 1)}+\vb^{(\ell)}_i|}{\|\vw^{(\ell)}_i\|}
\end{equation}
To measure the LC of the whole network we need to quantify distance for all neurons in all layers in the network as the input $\vx$ propagates through the network. If the distances for all the nonlinearity boundaries are larger, when the representations are located deep inside the linear regions of the network, resulting in a lower LC, while if the distances are smaller, the representations are at the edge of changing the preactivations and crossing a nonlinearity boundary, resulting in a higher LC.

\subsection{Bounds for Activation Magnitudes}
\label{app:ma-bounds}

Building on the discussion from previous sections, this section shows that massive activations if large enough guarantee low LC of the network around the data points. To satisfy this condition, we should be able to show that for any desirable distance $\delta>0$, there exists a massive activation magnitude such that all the distances between the representations and nonlinearity boundaries given in \eqref{eq:rep-dist} are larger than $\delta>0$. Proposition \ref{prop:ma-layer-hyp} provides the derivation for the activation magnitude required to migrate the regions in a single layer and Theorem \ref{thm:ma-full-net} extends the layerwise bound to the full network.

\SingleLayerMA*
\begin{proof}
For any neuron $i$ in layer $\ell$, the reverse triangle and
Cauchy--Schwarz inequalities give
\begin{align}
\operatorname{dist}(\vx^{(\ell-1)},\mathcal{H}_i^{(\ell)})
&=
\frac{|{\vw_i^{(\ell)}}^\top\vx^{(\ell-1)}+b_i^{(\ell)}|}
{\|\vw_i^{(\ell)}\|} \\
&\geq
\frac{
|[\vw_i^{(\ell)}]_k|\,|\vx_k^{(\ell-1)}|
-\|\vw_{i,-k}^{(\ell)}\|\,\|\vx_{-k}^{(\ell-1)}\|
-|b_i^{(\ell)}|
}{\|\vw_i^{(\ell)}\|} \\
&\geq
\frac{
|\vx_k^{(\ell-1)}|
\bigl(|[\vw_i^{(\ell)}]_k|
-\eta\|\vw_{i,-k}^{(\ell)}\|\bigr)
-|b_i^{(\ell)}|
}{\|\vw_i^{(\ell)}\|}.
\end{align}
Hence this lower bound exceeds
$\delta$ whenever
\begin{equation}
|\vx_k^{(\ell-1)}|
>
\frac{
\delta\|\vw_i^{(\ell)}\|+|b_i^{(\ell)}|
}{
|[\vw_i^{(\ell)}]_k|-\eta\|\vw_{i,-k}^{(\ell)}\|
}.
\end{equation}
Taking the maximum over all neurons $i$ concludes the proof.
\end{proof}

\FullNetworkMA*
\begin{proof}
Consider any point $\widetilde{\vx}^{(0)}$ satisfying
\begin{equation}
\|\widetilde{\vx}^{(0)}-\vx^{(0)}\|\leq\delta,
\end{equation}
and propagate it through the same network. Since ReLU is
$1$-Lipschitz, every residual layer satisfies
\begin{align}
\|\widetilde{\vx}^{(\ell)}-\vx^{(\ell)}\|
&\leq
\|\widetilde{\vx}^{(\ell-1)}-\vx^{(\ell-1)}\|
+
\|\mW^{(\ell)\top}
(\widetilde{\vx}^{(\ell-1)}-\vx^{(\ell-1)})\| \\
&\leq
(1+\alpha)
\|\widetilde{\vx}^{(\ell-1)}-\vx^{(\ell-1)}\|.
\end{align}
Applying this recursively gives
\begin{equation}
\|\widetilde{\vx}^{(\ell-1)}-\vx^{(\ell-1)}\|
\leq
(1+\alpha)^{\ell-1}\delta.
\end{equation}

Since the massive activation also persists through the residual stream and ReLU is nonnegative
\begin{equation}
\vx_k^{(\ell)}
\geq
\vx_k^{(\ell-1)}.
\end{equation}
Hence $\vx_k^{(\ell-1)}\geq\vx_k^{(0)}$ at every layer.

The lower bound on $\vx_k^{(0)}$ and Proposition~\ref{prop:ma-layer-hyp}, applied with distance $(1+\alpha)^{\ell-1}\delta$ give
\begin{equation}
\mathrm{dist}
\left(
\vx^{(\ell-1)},\mathcal{H}_i^{(\ell)}
\right)
>
(1+\alpha)^{\ell-1}\delta
\end{equation}
for every neuron $i$ in every layer $\ell$. Since every propagated point $\widetilde{\vx}^{(\ell-1)}$ lies within $(1+\alpha)^{\ell-1}\delta$ distance of $\vx^{(\ell-1)}$ it does not belong to $\mathcal{H}_i^{(\ell)}$ and because the initial point $\widetilde{\vx}^{(0)}$ was chosen arbitrarily, the propagated ball intersects no neuron hyperplane in the network.
\end{proof}

These results show that, with sufficiently large activations, the layer representations lie at least distance $\delta$ away from the neuron hyperplanes. This provides a sufficient condition for low LC. We note that the theoretical argument claims if the massive activation emerges at layer $\ell$ then that the local geometry is simplified for the subsequent layers of the network. As discussed in Section \ref{sec:ma-cause-rm}, in practice massive activations typically emerge in the first layer, so it is applied to layers and the subnetwork onwards from that point.

\begin{figure}[t]
  \centering
  \includegraphics[width=1.0\linewidth]{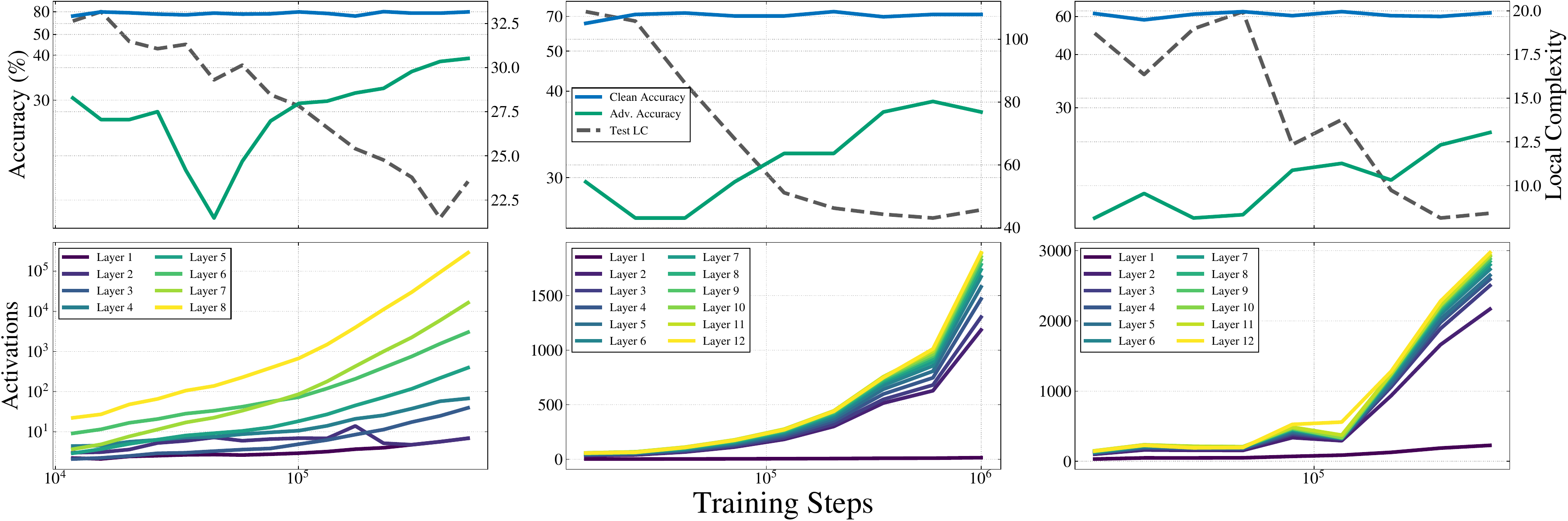}
  \caption{Emergence of the massive activations, region migration and adversarial robustness, in ResNet18 on CIFAR10 (left), ViT Tiny on CIFAR10 (middle) and ViT Small on Imagenette (right). Plots follow the same format as Figure~\ref{fig:mlp_gpt_pythia}. Adversarial radius of $\epsilon=0.0025$ is chosen in all evaluations.}
  \label{fig:resnet_vit}
\end{figure}

\section{Experimental Details}
\label{sec:exp_details}

\subsection{Training Details}

The following sections provide information about the optimizer and its parameters, loss functions, the setting and parameters used for measuring adversarial robustness and LC. We use the \texttt{PyTorch} deep learning library \citep{NEURIPS2019_9015} for all the experiments in this project.

\paragraph{Architectures.} We utilize ResNet18 architecture without Batch Normalization \citep{ioffe2015batch}, as it undoes delayed robustness \citep{humayun2024deep}, a pre-norm \citep{xiong2020on, radford2019language} Vision Transformers \citep{darcet2024vision} and pre-norm GPT based on the nanoGPT architecture \citep{Karpathy2022}. Unlike Batch Normalization we keep the Layer Normalization \citep{ba2016layer} operations in the transformer architectures including in the Pythia family and observe that it does not prevent delayed robustness.

\paragraph{Optimizer and Weight Decay.} Following \citet{humayun2024deep}, we use the Adam optimizer \citep{kingma2015adam} across all the experiments. \citet{liu2023omnigrok} claim that the delayed generalization or grokking in the network can be accelerated with weight decay, however \citet{humayun2024deep} found that in delayed robustness weight decay has the opposite effect and postpones the LC double descent phenomenon. Hence, we choose weight decay consistent with \citet{humayun2024deep} and unless specified otherwise use a weight decay of zero for all the experiments, except for the MLP architecture, where the weight decay is set to $0.01$. We use a learning rate of $0.001$ for MLPs, ResNets and GPTs and $3 \times 10^{-4}$ for ViTs. For MLPs and ResNets we use a batch size of $256$ and for ViTs and GPTs $64$ across all the experiments.

\paragraph{Loss Functions.} Following the experimental setup in \citet{humayun2024deep}, we use the mean squared error loss function for training our residual MLPs and the cross-entropy loss function for ResNets, ViTs and GPTs. No label smoothing is used for any of the experiments.


\subsection{Metrics}
\label{sec:metrics}

\paragraph{Adversarial Accuracy.} We quantify adversarial accuracy with AutoAttack \citep{croce2020reliable}, which is a union of APGD-CE and APGD-DLR \citep{croce2020reliable}, FAB \citep{croce2020minimally} and Square \citep{andriushchenko2020square} attacks. Unlike \citet{humayun2024deep} that uses only a PGD attack, our stronger attack scheme validates that the adversarial gains persist and are not caused by phenomenon like gradient masking \citep{carlini2019evaluating}. Adversarial accuracy is defined as the fraction of the data points for which the network maintains its correct prediction under AutoAttack. Unless specified otherwise we use $T=10$ steps and $5$ restarts for APGD-CE, APGD-DLR and FAB and $2000$ queries for Square attack. Square attack is only applied to image tasks. For MLP, ResNet18 and ViT the adversarial attack is run on the input images and for GPT and Pythia the adversarial accuracy is measured on the next token prediction task and perturbations are applied on the input token embeddings.

\paragraph{Local Complexity.} Following the LC metric hyperparameter choices in \citet{humayun2024deep} the cross-polytopal frame $\mV_\vx$ is constructed with $P=12$ orthonormal vectors and radius $r=0.005$ for MLPs, $r=0.001$ for ResNets and $r=0.005$ for ViTs, GPTs and Pythia across all the experiments. The LC is computed by counting the number of intersections of the embedded frame with the nonlinearity boundaries.

\paragraph{Massive Activations Magnitudes.} 
We track the sum of the top-$3$ activations in the residual stream across all the layers for all architectures. For ViTs we track the massive activations on the CLS token only, while for GPT and Pythia the activations are tracked in all token representations and the averaged value is reported.

\begin{figure}[t]
    \begin{minipage}{0.24\textwidth}
        \centering{\includegraphics[width=\linewidth]{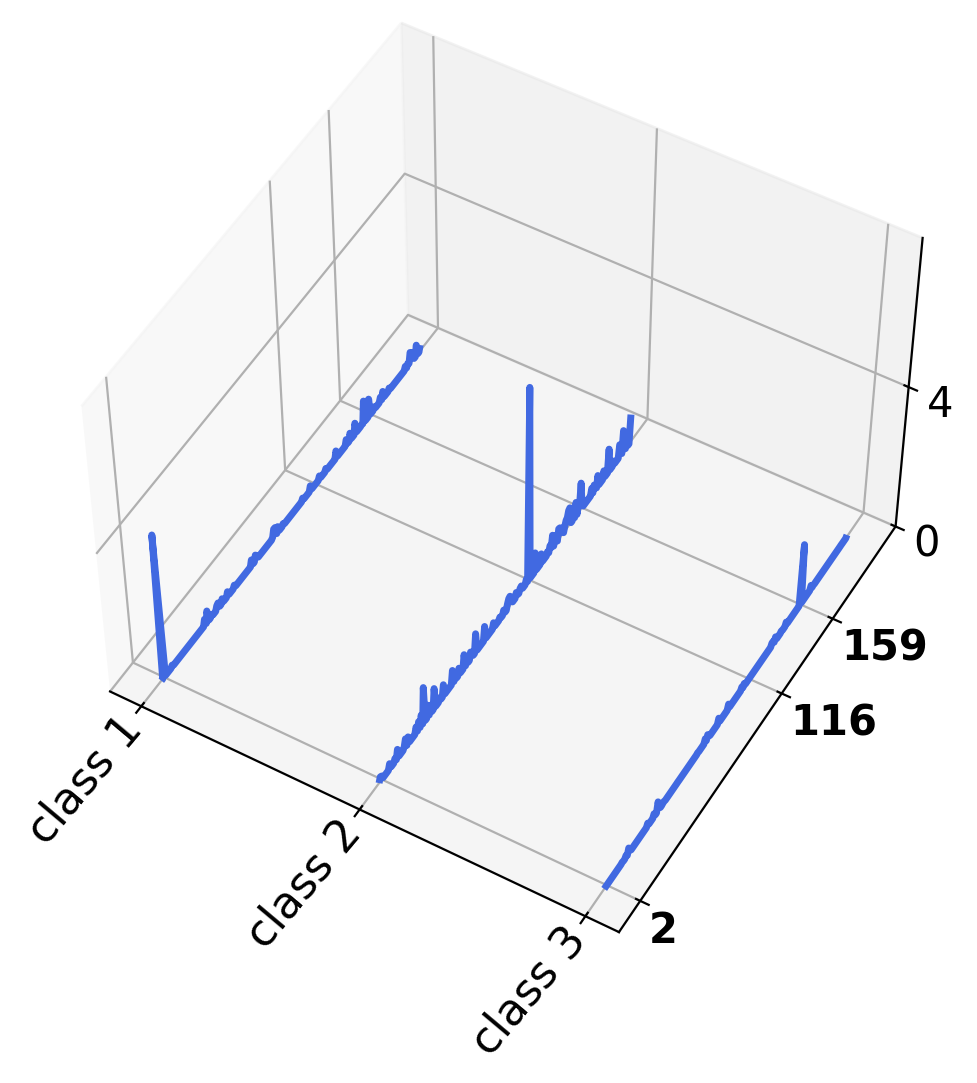}}
    \end{minipage}
    \begin{minipage}{0.24\textwidth}
        \centering{\includegraphics[width=\linewidth]{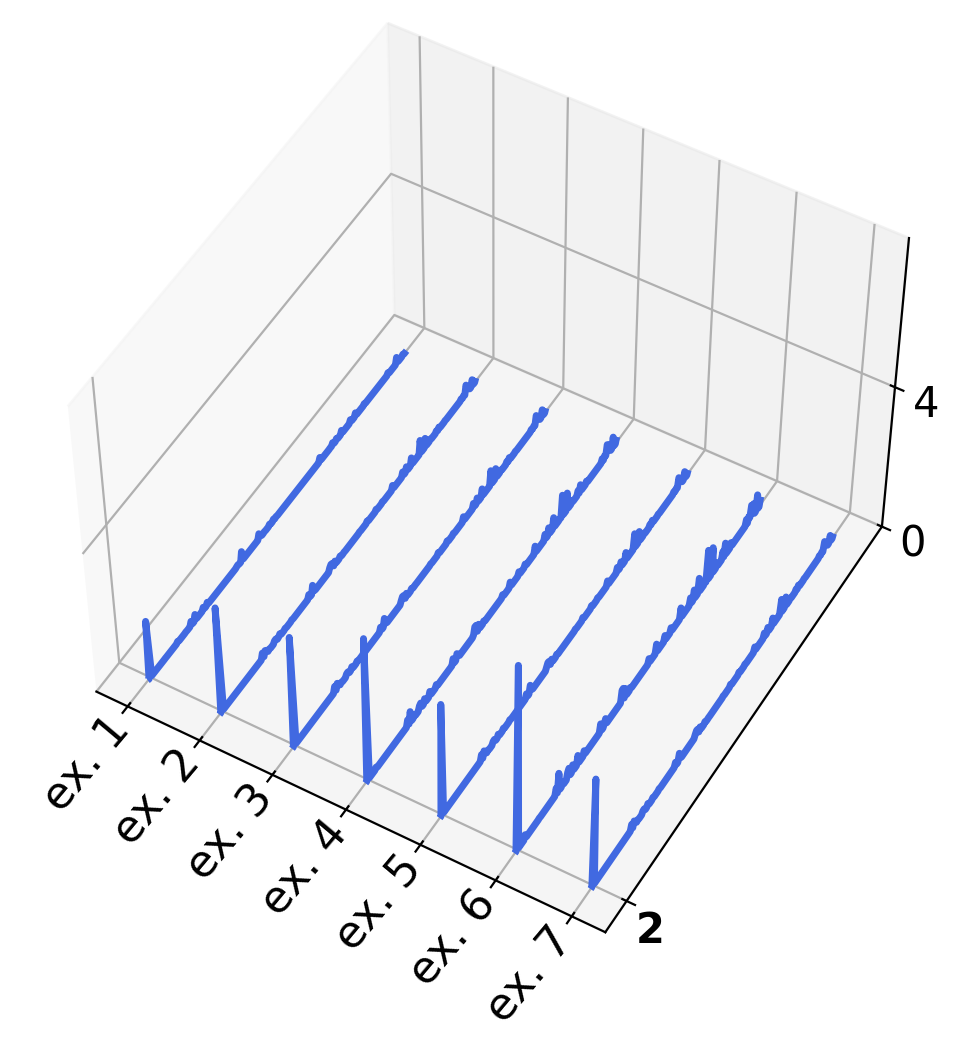}}
    \end{minipage}
    \begin{minipage}{0.25\textwidth}
        \centering{\includegraphics[width=\linewidth]{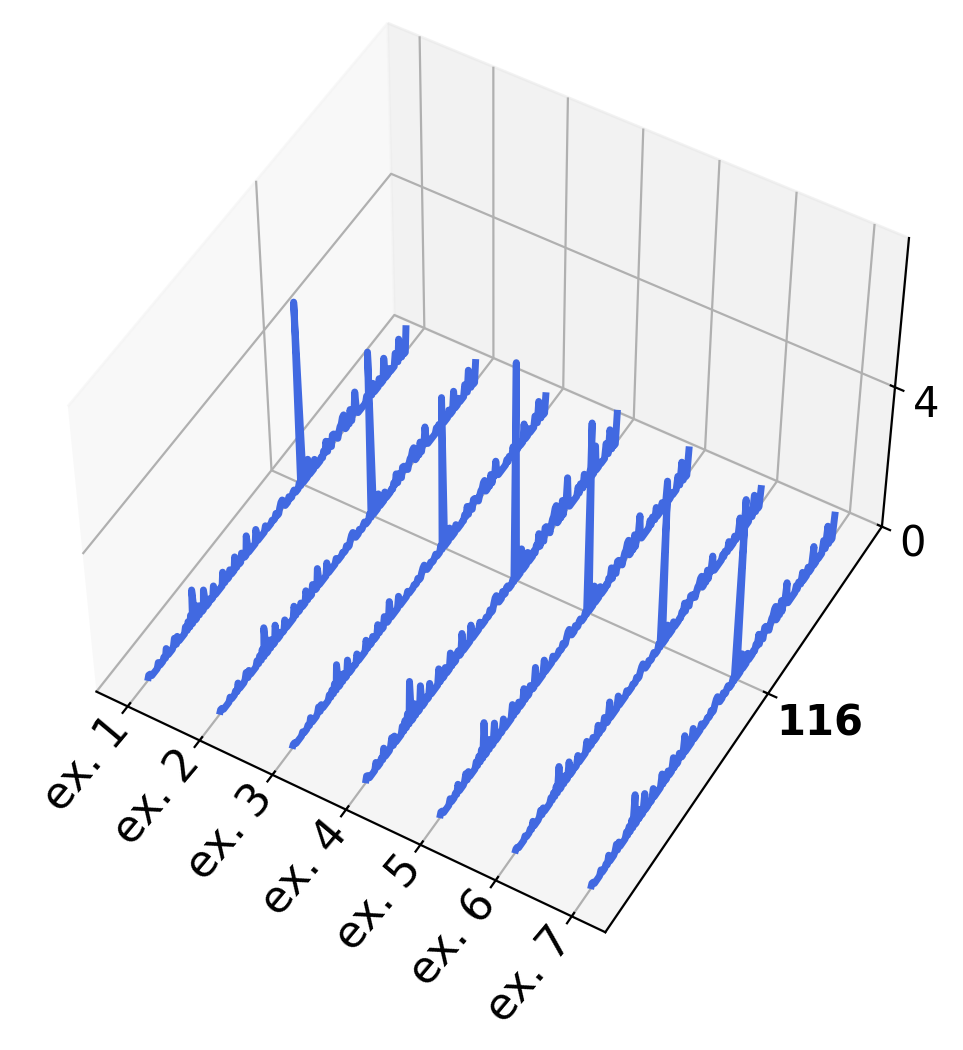}}
    \end{minipage}
    \begin{minipage}{0.24\textwidth}
        \centering{\includegraphics[width=\linewidth]{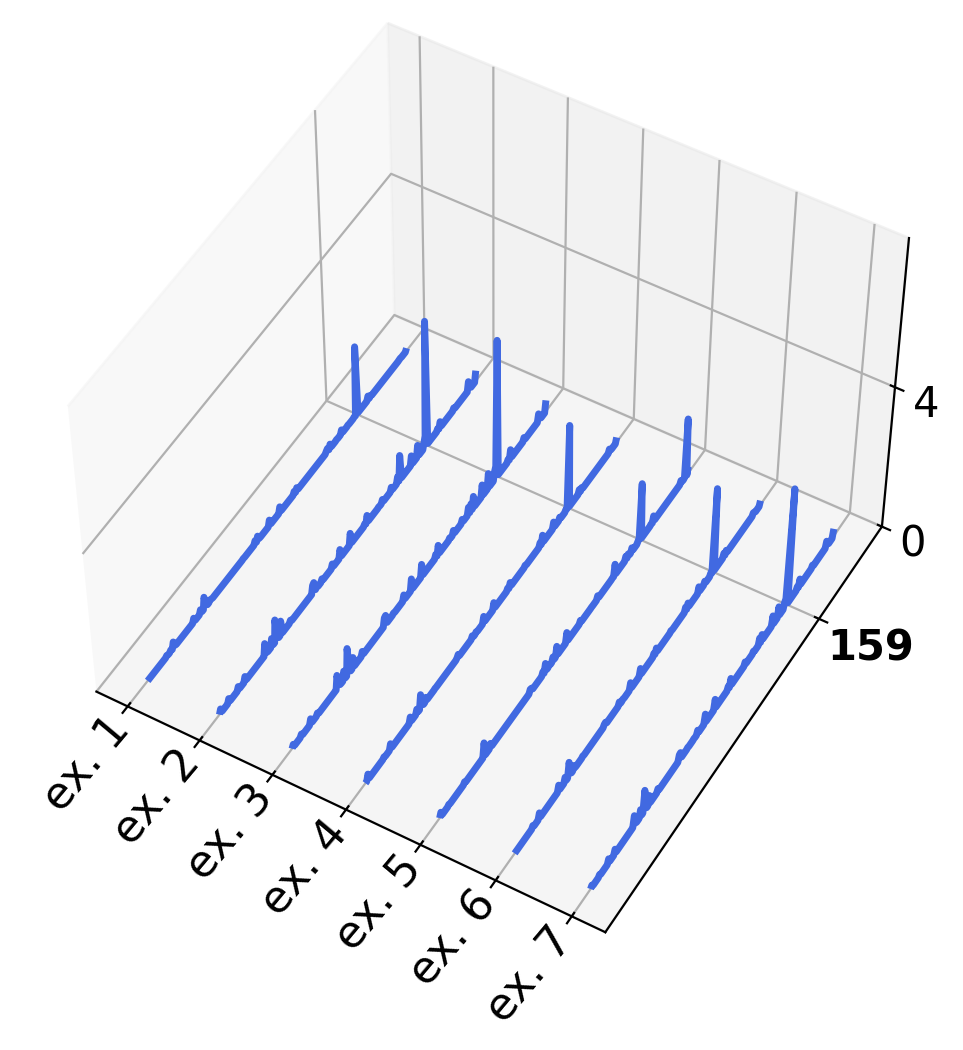}}
    \end{minipage}
    \caption{The activations have different patterns for different classes and activate on the same positions for samples within the same class. This figure represents the activation patterns prior to entering the last layer for the $4$ layer residual MLP with width $200$ trained on MNIST. \textbf{Left} figure shows that the activation patterns are different from three distinct classes in MNIST. \textbf{Middle-left}, \textbf{middle-right} and \textbf{right} figures show the activation patterns of seven examples from the first, second and third classes outlined on the left plot respectively, showing that the indices where the large activations appear are consistent.}
    \label{fig:mlp-class-ex-ma}
\end{figure}

\section{Additional Experiments}

\subsection{Massive Activations and Delayed Robustness}
\label{sec:app-ma-dr}

We provide additional experiments on the emergence of massive activations and delayed robustness in ResNet18 on CIFAR10, ViT Tiny on CIFAR10 and ViT Small on Imagenette illustrated in Figure \ref{fig:resnet_vit}.

\subsection{Massive Activation Patterns}

Different classes exhibit different massive activation patterns, while samples from the same class have activations in the same positions as depicted in Figure \ref{fig:mlp-class-ex-ma}. This indicates that the network sends different classes into separate parts of the space, that are free of nonlinearities. As a result, the nonlinearity boundaries accumulate in the part of the space, which separates the different classes, consistent with the migration towards decision boundary observation in \citet{humayun2024deep} as illustrated in Figure \ref{fig:mlp_acc_lc_splinecam} (right).

\subsection{Suppressing Massive Activations}
\label{sec:sup-ma-rm}

\begin{wrapfigure}{r}{0.40\textwidth}
    \centering
    \vspace{-3.0em}\includegraphics[width=1.0\linewidth]{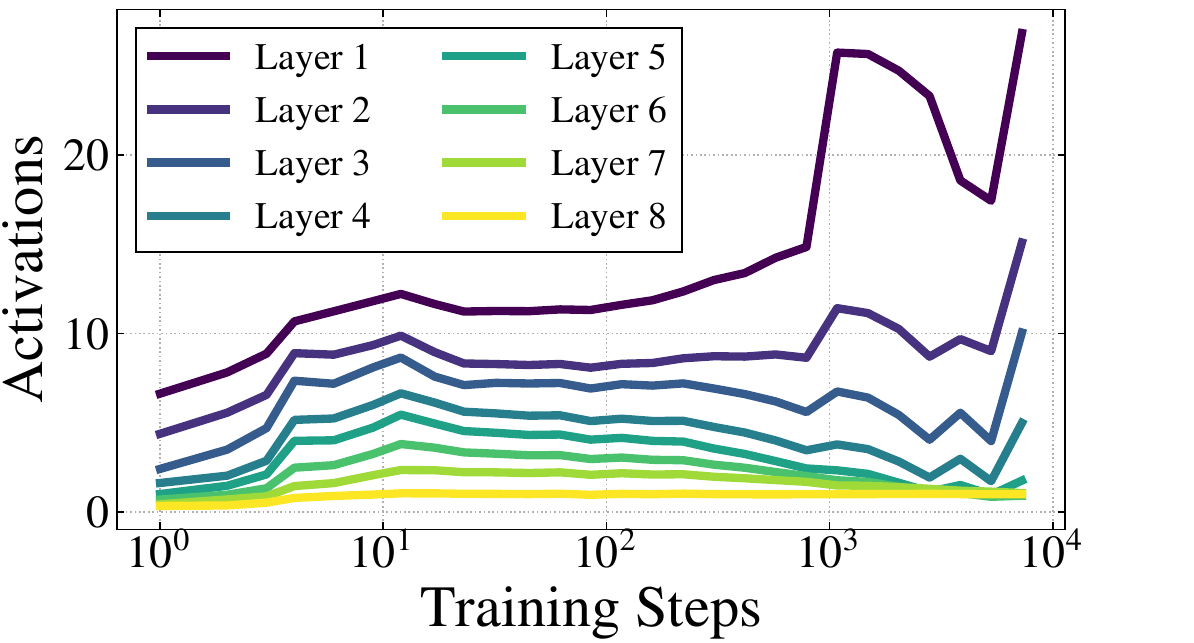}
    \caption{MLP without residual connections on MNIST. Massive activations emerge in the first layer, but gradually fade away with each layer unlike in residual MLP, where they are propagated through the layers via the residual stream.}
    \label{fig:mlp-no-res}
    \vspace{-1.5em}
\end{wrapfigure}

\begin{figure}[h]
\centering
\begin{minipage}{1.0\textwidth}
  \centering{\includegraphics[width=\linewidth]{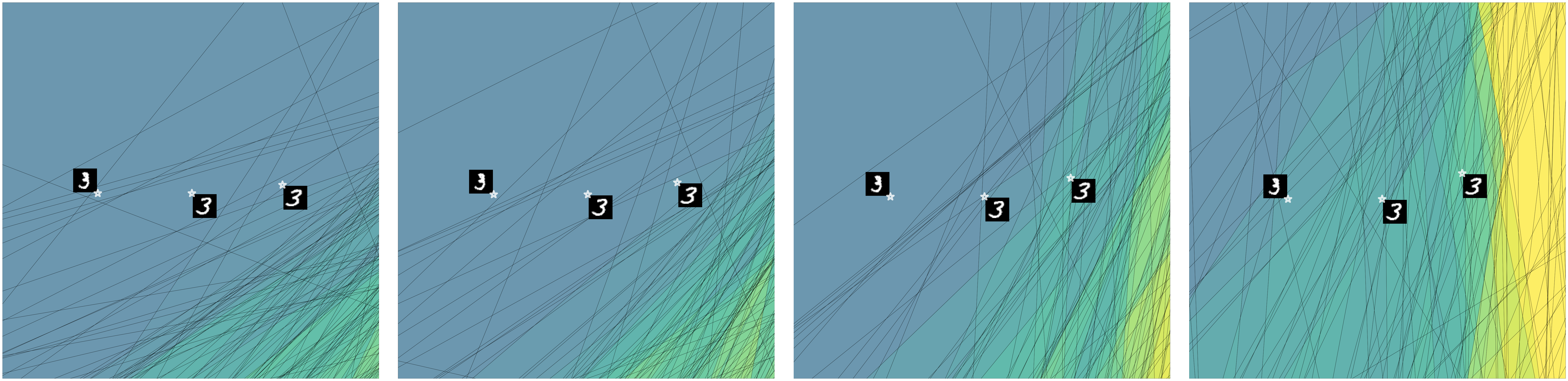}}
\end{minipage}

\vspace{0.26em}

\begin{minipage}{1.0\textwidth}
    \centering{\includegraphics[width=\linewidth]{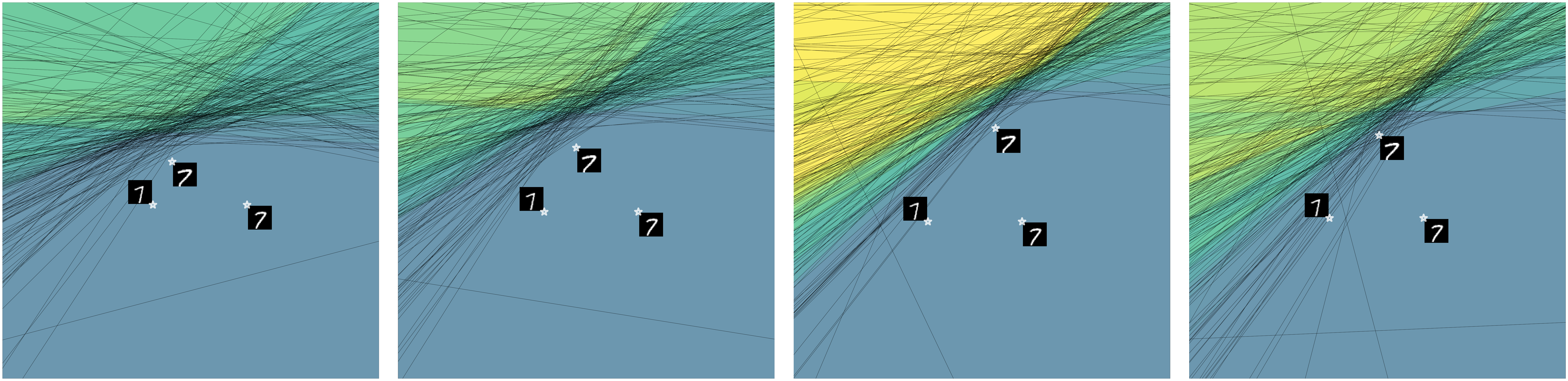}}
\end{minipage}

\vspace{0.26em}

\begin{minipage}{1.0\textwidth}
    \centering{\includegraphics[width=\linewidth]{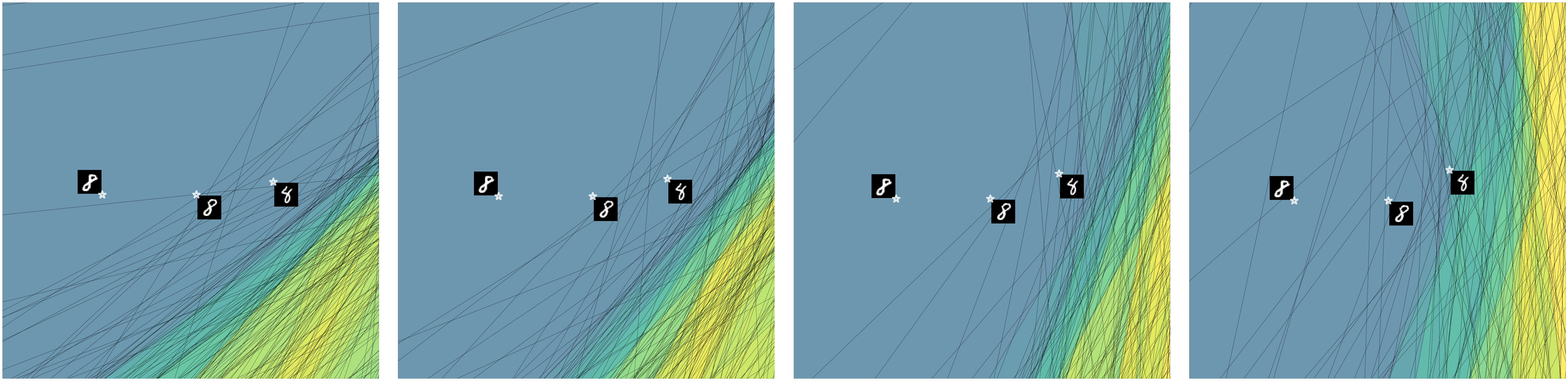}}
\end{minipage}
\caption{SplineCam visualization of the input space partitioning of the last layer of a depth $4$ and $200$ width residual MLP trained on MNIST as the dominant activations in the input are suppressed from left to right. \textbf{Left} represents the partitioning without any input modifications, serving as a reference, \textbf{right} with the dominant activation set to zero, \textbf{middle left} and \textbf{middle right} with the dominant activation set to $66\%$ and $33\%$ of its original value respectively.}
\label{fig:sup-ma-more}
\end{figure}

This section supplements Section \ref{sec:supp-ma}, providing more examples of the input space geometry when suppressing the dominant activations. Figure \ref{fig:sup-ma-more} shows the SplineCam visualizations of an MLP on MNIST for different classes, as the dominant activations are manually suppressed. We consistently observe that reducing these activations magnitudes pushes the input representations back to the regions where nonlinearity boundaries are densely accumulated. Importantly, the consequences get more severe with more suppression of the dominant activations as exemplified in Figure \ref{fig:sup-ma-more}.



\subsection{Massive Activations in Non-Residual MLPs}
\label{sec:ma-non-res-nets}

This section studies the massive activation dynamics in MLP without residual connections. We use $8$ layer and $200$ width MLP and train it on MNIST and track the accuracy, adversarial accuracy, LC and activation magnitudes according to the methodology described in Appendix \ref{sec:exp_details}. Figure \ref{fig:mlp-no-res} shows the result of the experiment, where massive activations emerge in the first layer however they slowly fade away in the following layers, confirming that residual stream is the primary enabler for the propagation of massive activations through the network.

\subsection{Massive Activations in Qwen 2.5 and Gemma 3}
\label{sec:ma-qwen-gemma}

Figure \ref{fig:qwen-suite-ma} and \ref{fig:gemma-suite-ma} show the massive activations in the hidden representations of Qwen 2.5 and Gemma 3 model families across different sizes and similar to the Pythia suite they reside in each token representation. Additionally we observe that the activation magnitudes increase with the size of the model for each model family, which matches the trend in the Pythia suite, presented in Figure \ref{fig:pythia-suite-ma}.


\subsection{Massive Activation Magnitudes}

\begin{figure}[t]
  \begin{minipage}{0.24\textwidth}
      \centering{\includegraphics[width=\linewidth]{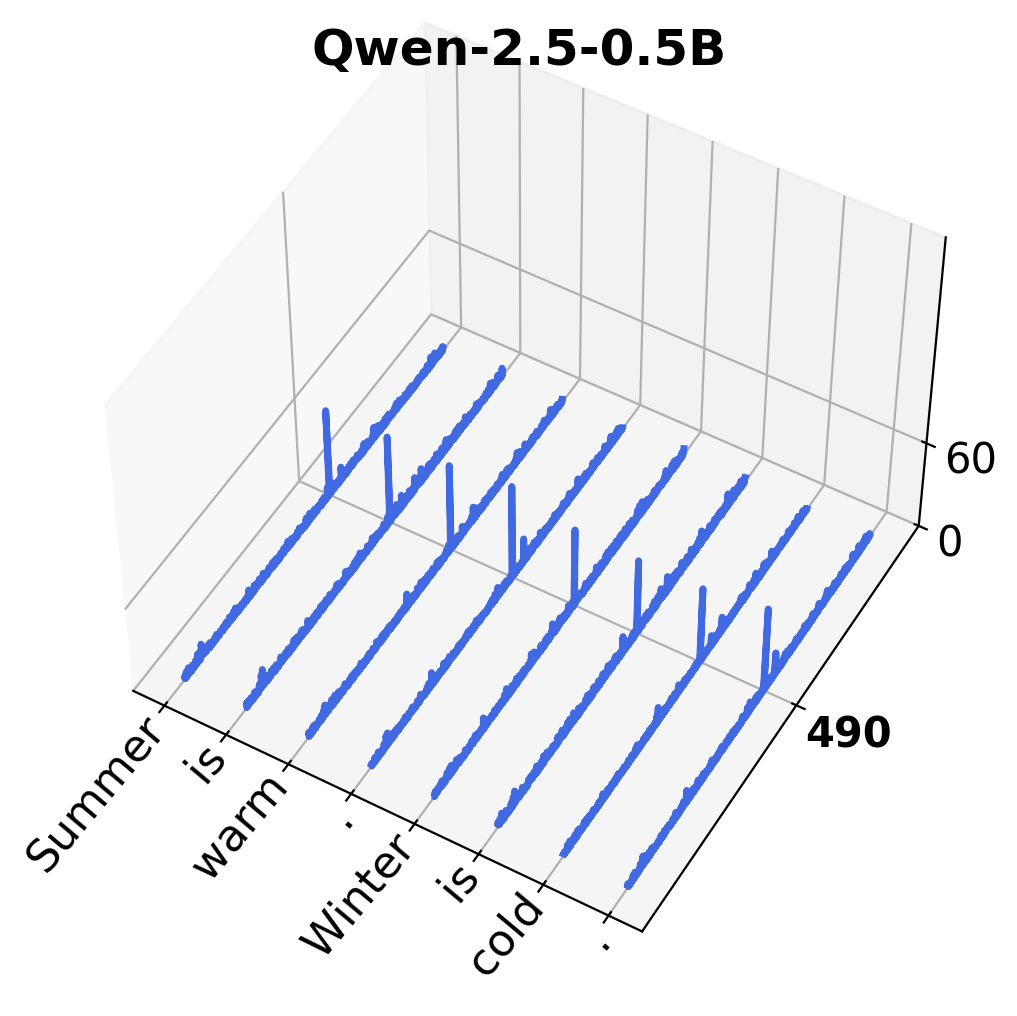}}
  \end{minipage}
  \begin{minipage}{0.24\textwidth}
      \centering{\includegraphics[width=\linewidth]{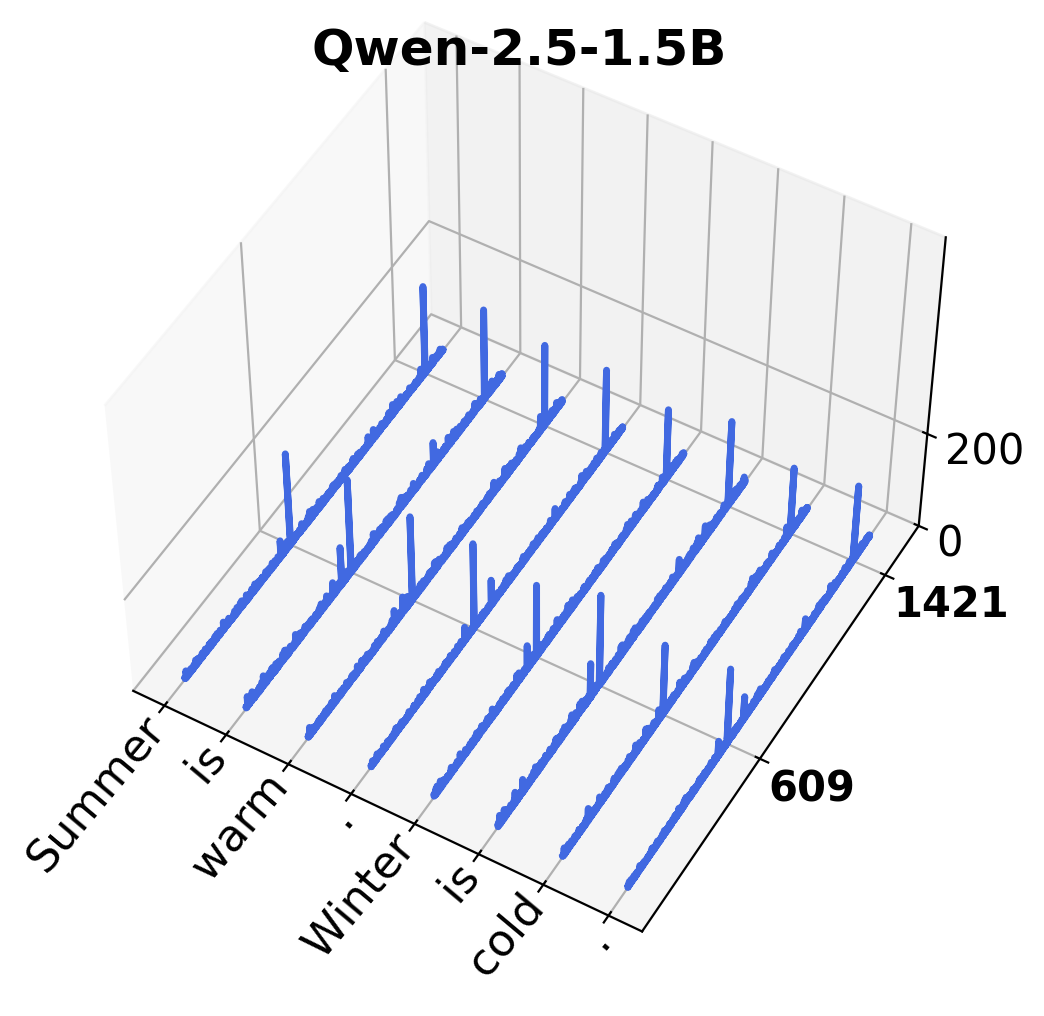}}
  \end{minipage}
  \begin{minipage}{0.25\textwidth}
      \centering{\includegraphics[width=\linewidth]{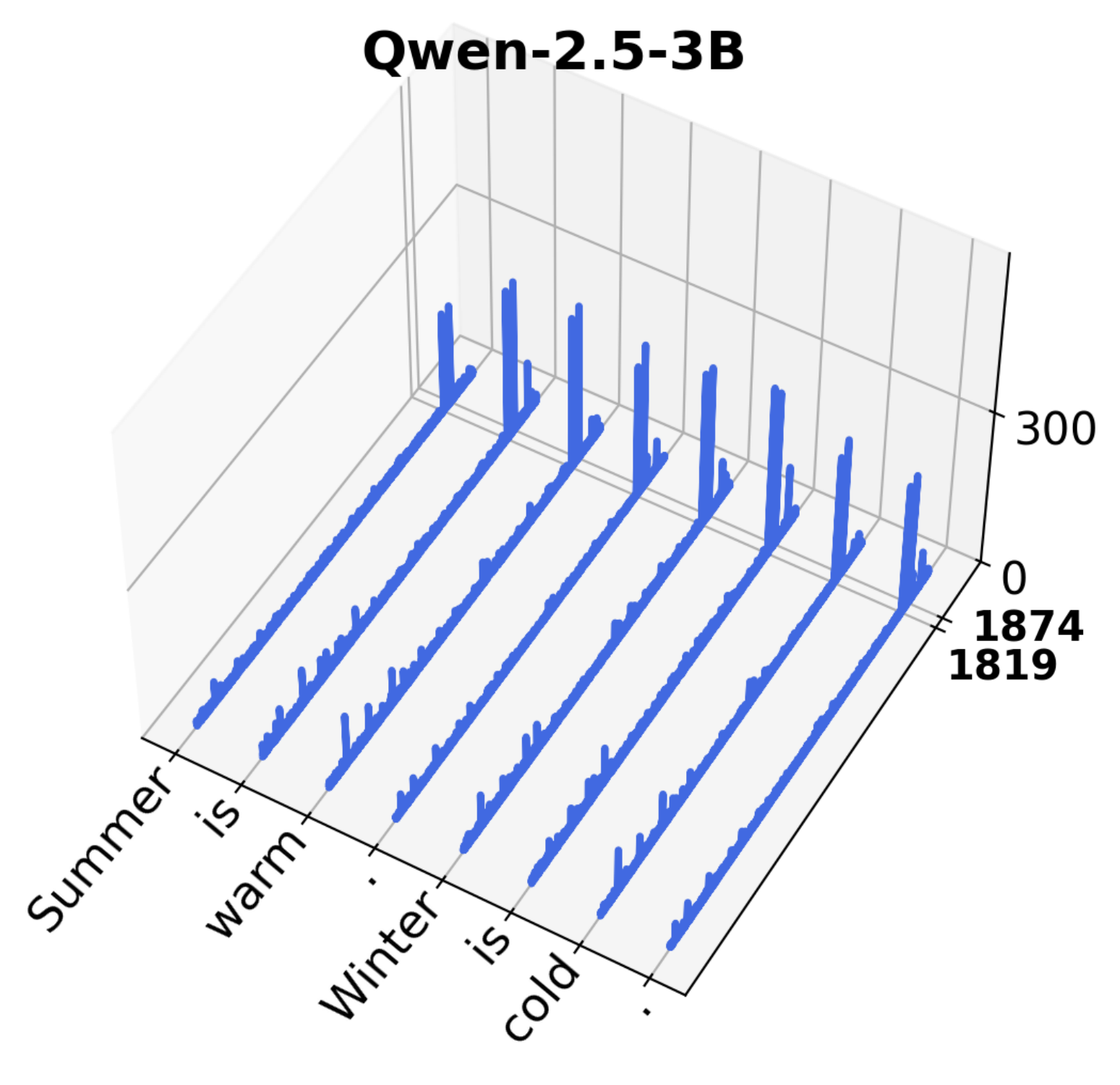}}
  \end{minipage}
  \begin{minipage}{0.24\textwidth}
      \centering{\includegraphics[width=\linewidth]{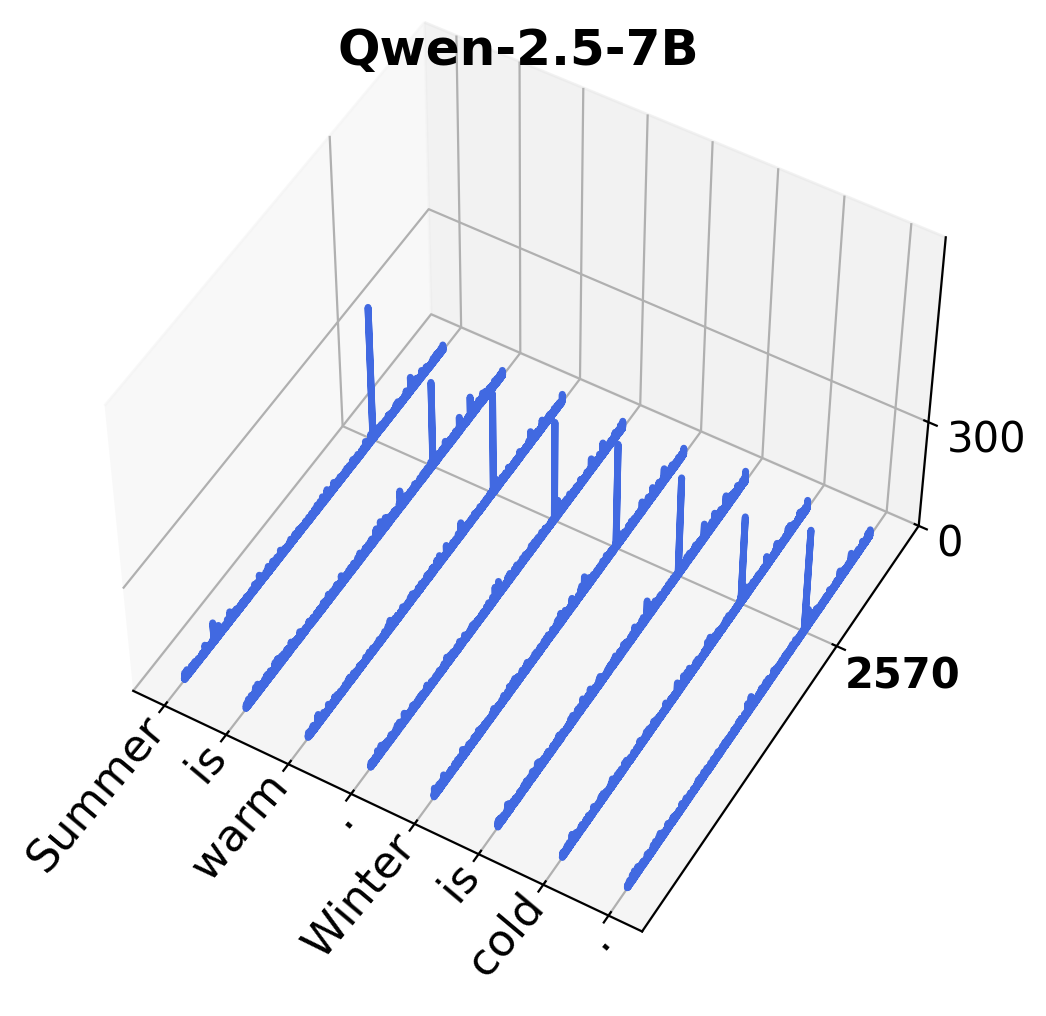}}
  \end{minipage}
  \caption{Massive activations in Qwen 2.5 family of pretrained models across different sizes.}
  \label{fig:qwen-suite-ma}
\end{figure}

\begin{figure}[t]
  \begin{minipage}{0.24\textwidth}
      \centering{\includegraphics[width=\linewidth]{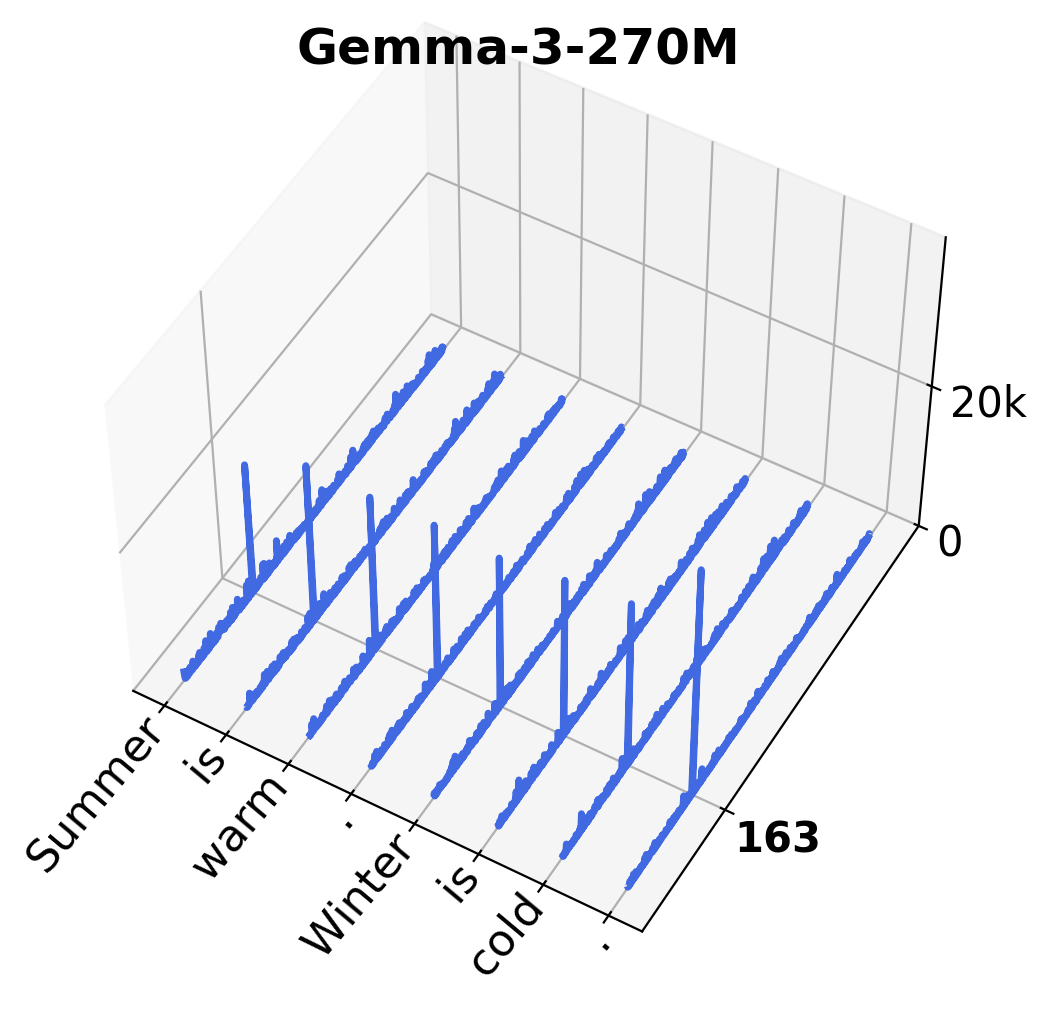}}
  \end{minipage}
  \begin{minipage}{0.24\textwidth}
      \centering{\includegraphics[width=\linewidth]{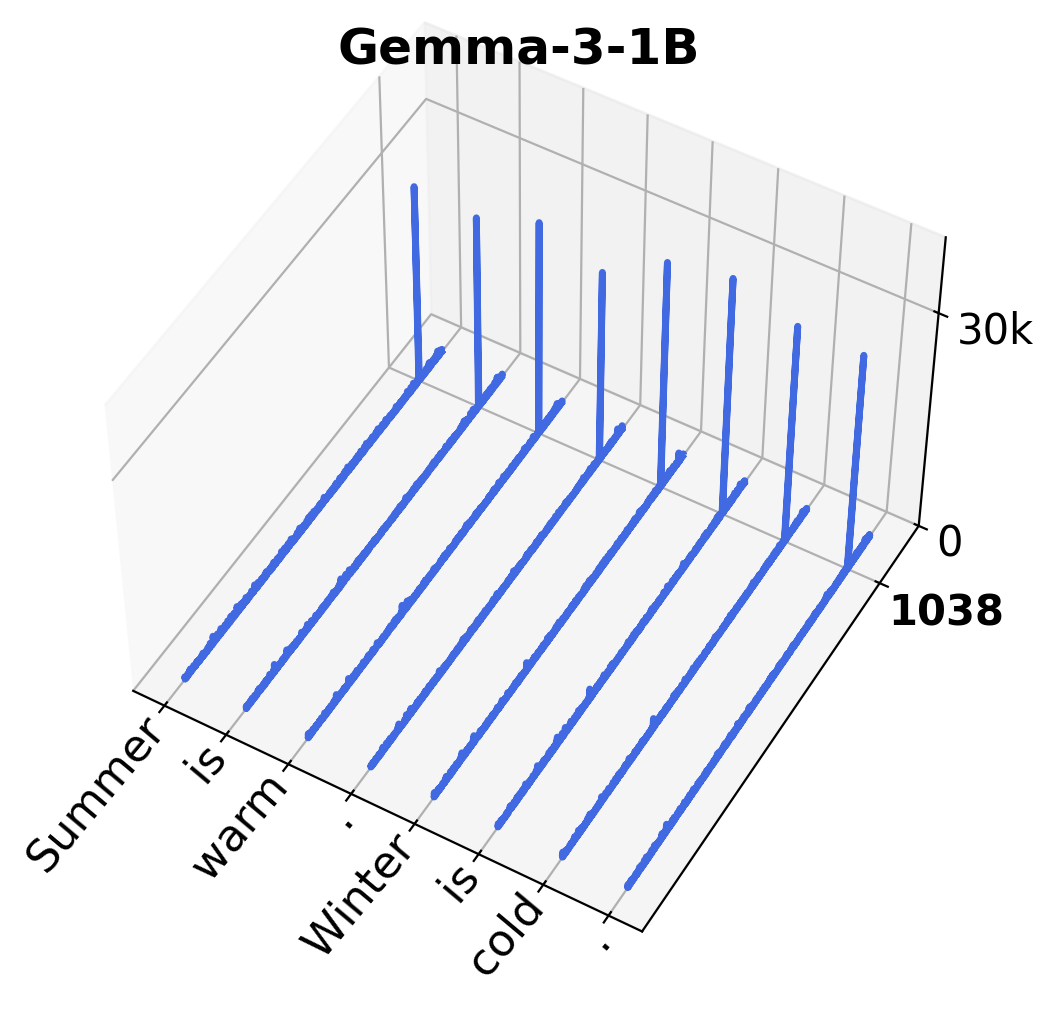}}
  \end{minipage}
  \begin{minipage}{0.25\textwidth}
      \centering{\includegraphics[width=\linewidth]{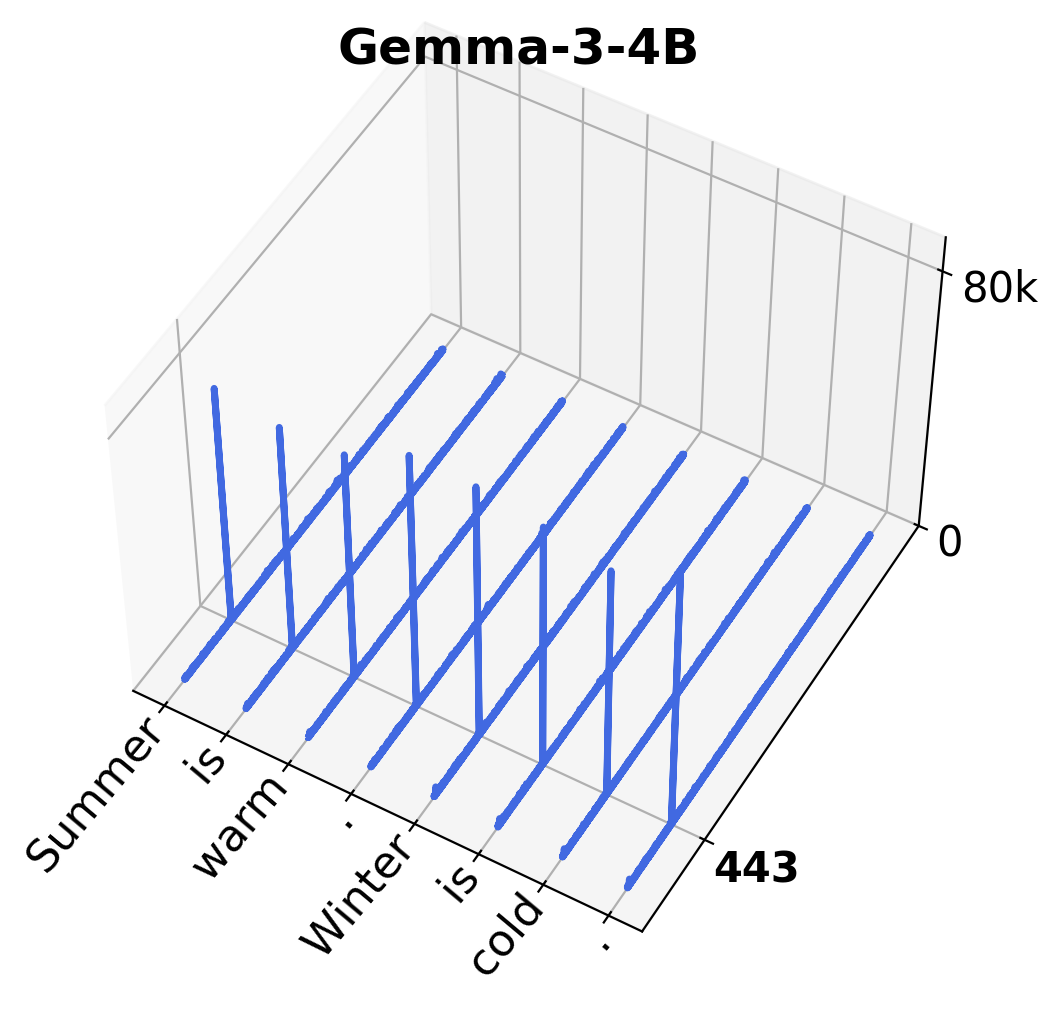}}
  \end{minipage}
  \begin{minipage}{0.24\textwidth}
      \centering{\includegraphics[width=\linewidth]{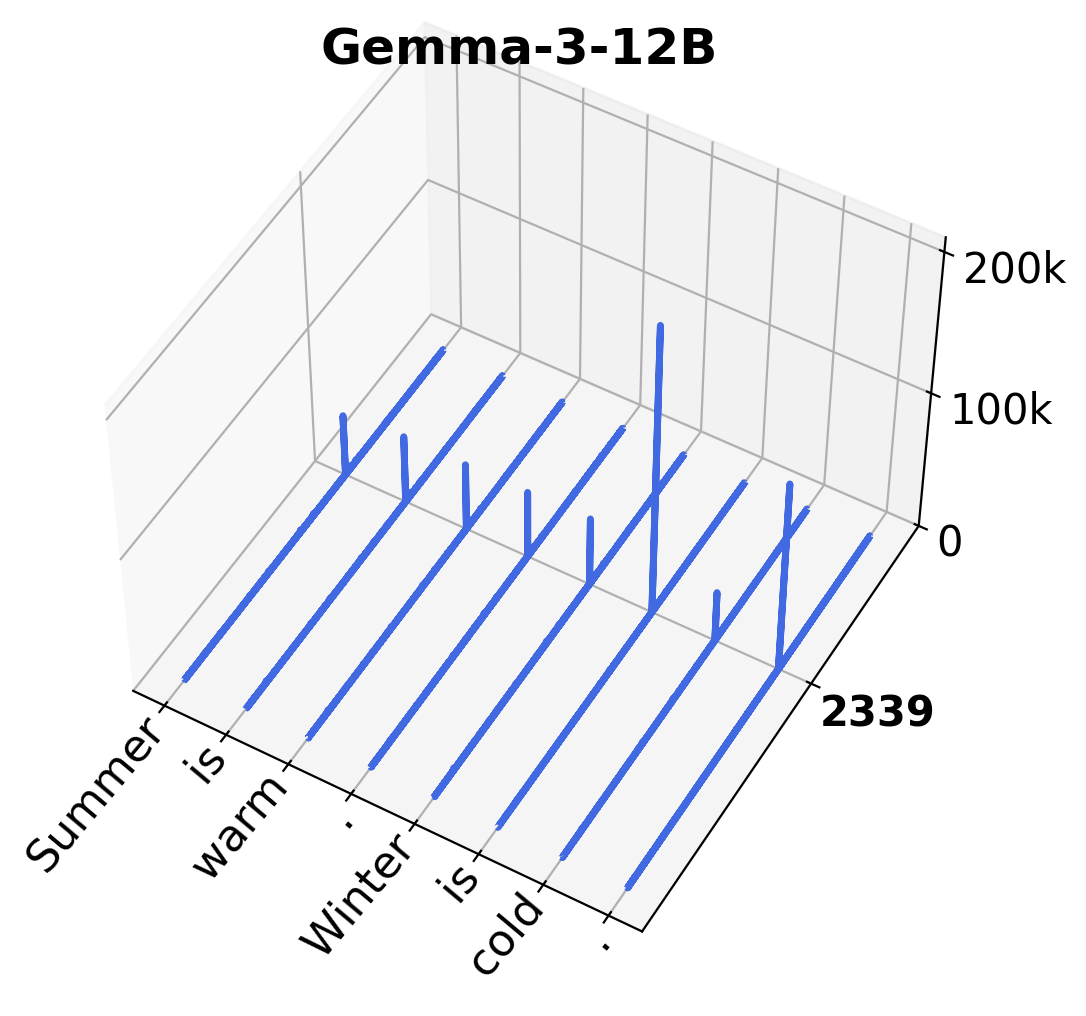}}
  \end{minipage}
  \caption{Massive activations in Gemma 3 family of pretrained models across different sizes.}
  \label{fig:gemma-suite-ma}
\end{figure}

We highlight that massive activation magnitudes differ in different architectures. Figure \ref{fig:resnet-vit-gpt-ma} shows that they are much larger in ResNet18 ($\approx15$k) compared to ViT Small ($\approx2$k) and GPT ($\approx6$k). Additionally, we notice this trend in the Pythia, Qwen 2.5 and Gemma 3 LLMs as depicted in Figures \ref{fig:pythia-suite-ma}, \ref{fig:qwen-suite-ma} and \ref{fig:gemma-suite-ma}, where the magnitudes of the largest activations increase with the scale of the model. We conjecture that these magnitudes have a dependence on the architectural configurations of the network, such as the width and depth. We believe that these questions can be explored in future works with more extensive empirical and theoretical developments.




\newpage

\end{document}

%% file: math_commands.tex
\usepackage{amsmath,amsfonts,bm}

\def\eqref#1{equation~\ref{#1}}

\def\1{\bm{1}}

\def\vb{{\bm{b}}}

\def\vv{{\bm{v}}}
\def\vw{{\bm{w}}}
\def\vx{{\bm{x}}}

\def\vz{{\bm{z}}}

\def\mV{{\bm{V}}}
\def\mW{{\bm{W}}}

\DeclareMathAlphabet{\mathsfit}{\encodingdefault}{\sfdefault}{m}{sl}
\SetMathAlphabet{\mathsfit}{bold}{\encodingdefault}{\sfdefault}{bx}{n}

\def\gV{{\mathcal{V}}}

